# Design and fabrication of autonomous electronic lablets for chemical control.

John S. McCaskill[1,4,5], Thomas Maeke[2], Dominic Funke[2], Pierre Mayr[2], Abhishek Sharma[1], Patrick F. Wagler[1], Jürgen Oehm[3]

[1]*Microsystems Chemistry and Biomolecular Information Processing (BioMIP), Ruhr University Bochum, Germany*
[2] *Integrated Systems (IS), Ruhr University Bochum, Germany*
[3] *Analog Integrated Circuits (AIS), Ruhr University Bochum, Germany*
[4] *Research Center for Materials, Architectures, and Integration of Nanomembranes (MAIN, Chemnitz University of Technology, 09126 Chemnitz, Germany*
[5] *European Centre for Living Technology (ECLT), Ca' Bottacin, Dorsoduro 3911, Venice, 30123, Italy.*

## ABSTRACT

*Lablets* are autonomous microscopic particles with programmable CMOS electronics that can control electrokinetic phenomena and electrochemical reactions in solution via actuator and sensor microelectrodes. The *lablets* are designed to be rechargeable using an integrated supercapacitor, and to allow docking to one another or to a smart surface for interchange of energy, electronic information and chemicals. In this paper, we describe the design and fabrication of singulated *lablets* (CMOS2) at the scale of 100 by 200 µm, with the supercap adjacent to the functional *lablet* and occupying half the space. In other works, we have characterized the supercap and described the electronic design and proven functionality using arrays of these *lablets*. Here we present fabrication details for integrating functional coatings and the supercap and demonstrate electronic functionality of the *lablets* following singulation.

# 1. Introduction

The programmable investigation and control of chemical systems at the microscale has been an increasingly successful area in microsystem technology for over 25 years including our own work in lab-on-a-chip and microfluidics to approach electronic chemical cells [1-2]. These systems require and are limited by their physical connection (wires, tubes, pipetting) to the macroscopic control system, both for electrical and chemical interfacing. Wireless electronic systems, communicating using radio waves, although already advocated for smart dust [3-4] and implemented down to mm scales, are not yet effective at 100 µm scales and below, especially in aqueous solution where communication is damped, and also do not provide a solution for powering smart microscopic electronic particles in solution. Our approach is a novel and more chemically inspired one [5] – to take advantage of the mobility of microscopic particles which allows their docking to one another pairwise or to a smart microstructured surface (called the dock). It involves fully programmable CMOS electronic particles in contrast to other more restricted approaches such as plasmonic smart dust [6].

Electronic integration using CMOS has been optimized for high speed (GHz range) operation and high integration levels with feature sizes down to 30nm and below. However, for microscopic electronics, extremely low power operation is required (total average power, typically ≤ 1 nW for 1000s) by current microscopic charge storage limitations (≤2 µF using supercap technology), which is not consistent either with high frequency operation or the leakage currents associated with the finest scale transistors. Instead, low power operation has been achieved using 180nm technology and an especially designed slow clock [7] and custom transistor design. Electronic actuation of chemical reactions mostly requires switching of voltages on microelectrodes in aqueous solution, which typically have significant capacitances, as exploited in electrolyte capacitors. Typical double layer and Stern capacitance values of 100pF for a 20 µm scale microelectrode in several mM salt solutions), with switching on a timescale of several µs. The fast switching of such electrodes can deplete the voltage on microscopic batteries, which would reset CMOS electronics. At 10 Hz, one such electrode, or at 1 Hz 10 electrodes can consume the entire available power. Supercapacitors have been proposed in other contexts to provide higher bursts of power and we have characterized a supercapacitor coating on arrays of CMOS lablet micro particles in separate work [8,9], for subsequent integration into singulated lablets, as will be described here.

In this work, we present progress in wafer scale manufacturing of lablets, focussing on the issues associated with integration of encapsulated supercaps and the coating of supercapacitors electrodes, and of actuators and sensors on the lablets. The electronic design and functionality of the CMOS2 lablets has been described and characterized in prior work on the lablet arrays [10], without the full processing required for complete supercap integration and microstructured surfaces. The work begins with the lablet physical design in section 2, presents the fabrication process in section 3 including the programmable galvanics used for coating and then proceeds to lablet characterization in section 4 and tests of functionality in section 5. The paper concludes with a summary of lessons learnt, the potential of this approach and strategies for improved manufacturing.

## 2. Lablet Physical Design

The overall considerations for lablet functional design has been described previously [5]: including the need for microelectrode actuators and sensors, a microstructured surface profile to support docking, limit chemical diffusion and control electrokinetic microfluidics, an integrated encapsulated supercap to support autonomous power, and microelectrodes for charging and communication.

Lablet particles were originally conceived with the dimensions 100x100x50µm so that two lablets could dock in solution, separated only by a sub-micrometer thin film, to form a cube with dimension 100µm. Thus the area available to CMOS electronics is at most 100x100 µm unless 3D integration can be employed. The first lablets designed and built (termed CMOS1 lablets) had these dimensions and their physical design and characterization is reported in the paper dealing with bipolar power [11]. The initially proposed layout for the second generation (CMOS2) lablets is shown in **Fig. 1**.

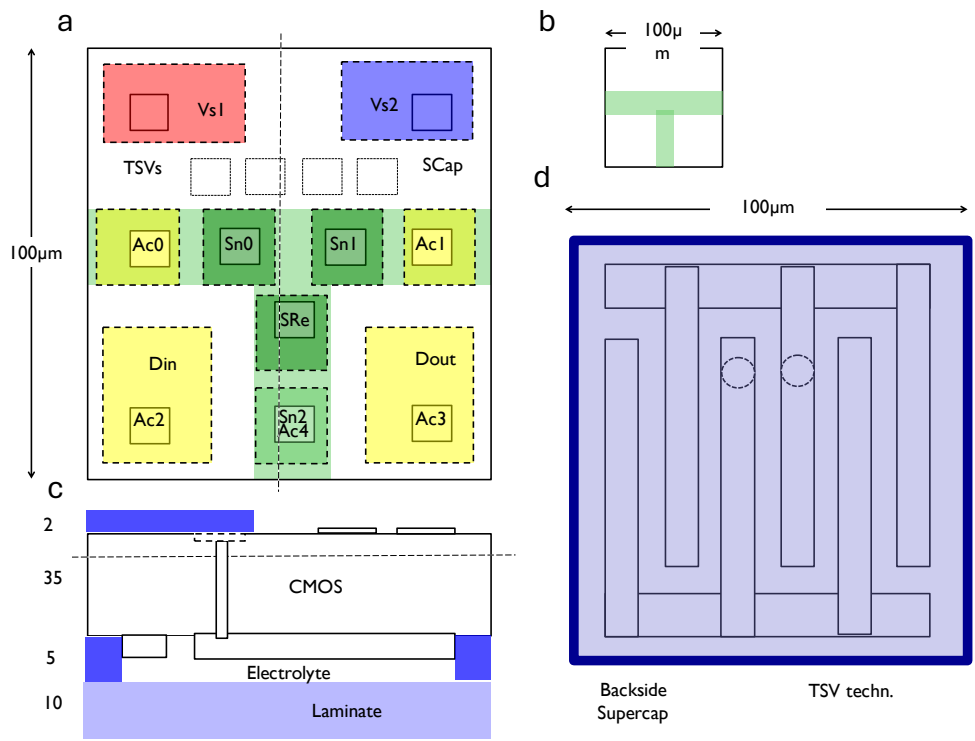

**Figure 1: Initial target structural design of CMOS2 lablets**. **a.** active surface view, showing recessed channel (green) with actuator (Ac) and sensor (Sn) electrodes, large charging (red/purple) and communication (yellow) electrodes and location of TSVs (see text) for supercap **b.** simplified view of T-shaped recessed channel on active surface **c.** cross-sectional view through the dashed line in a, showing the supercap on the backside with laminate-encapsulated electrolyte **d.** back side view of interdigital layout of electrodes in the supercap. This geometry was converted to a coplanar design without TSVs, see Fig. 2.

The intended supercapacitor integration on the back side (that opposing the active CMOS electronics) of the 2$^{nd}$ generation (CMOS2) lablets, reported here, would have required the integration of TSV (though silicon vias) to connect the supercap to front side electronics and double-sided thinned wafer processing, and while technically feasible in principle, access to these 3D connection processes introduced additional constraints in the overall fabrication which were complicating and delaying fabrication, and hence the decision was made to relegate these to a future generation of lablet fabrication. The supercap requires a surface area close to 100x100μm to deliver the required 1-2 μF capacitance in an interdigital structure (corresponding to a requisite supercapacitor coating performance of 40mF/cm$^2$). For this generation of lablets, the supercap was placed compactly, directly adjacent to the active surface, resulting in an intermediate revised lablet size target of 100x200x50μm.

Two geometric layouts of the constructed lablets for lateral and facial docking are shown in **Fig. 2**. The presence of a supercap on the same side of the lablet as the active surface makes docking to a planar surface impossible, unless the actuator and sensor structures are elevated to the top plane of lamination to allow proximate communication. A high aspect ratio galvanic technique (Temicon GmbH) for raising conducting structures had been investigated previously in test structures, but the achievement of sufficiently planar elevated structures for controlled docking was deemed unlikely. Instead, lablet-lablet docking can still be achieved facially with staggered binding where the two supercap structures do not oppose one another. For this docking, the optional thicker walls and laminate are removed from the active lablet surface (shown as optional structures in Fig. 2). Making use of the laminate to form enclosed channels above the active surface of the lablet (in addition to the supercap), docking to other lablets can still be achieved laterally on a surface by aligning channel entrances. Lablet docking to a smart surface can then only be achieved either face down with a 17μm gap between surface and lablet electrodes or face up (allowing direct observation) with a 30μm gap.

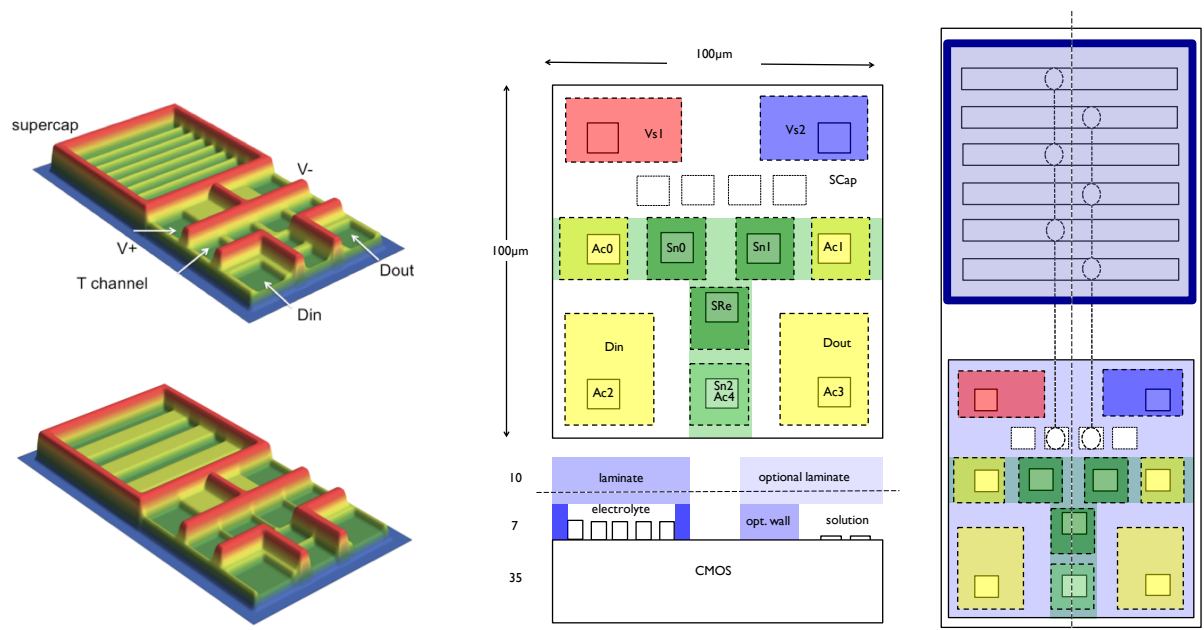

**Figure 2: CMOS2 lablets with coplanar supercap.** As described in Fig. 1, but here with the supercap structure on the same front side as CMOS and lablet actuators and sensors. **Left**: Computed height profile in two variants with different sized electrodes. The 3D geometry is computed from the masks, the height profile is exaggerated for clarity (height differences 7μm). **Right:** The design is shown in two variants with (i) optional walls and laminate for lateral lablet docking and communication via lateral channel openings (ii) open actuator and sensor electrodes in shallow channels for facial docking (to other lablets or a smart docking surface).

The widths of channel and supercap retaining walls are dimensioned as close to 10μm to allow reliable fluid seals. Charging and communication electrodes are dimensioned as large as possible to ensure effective function. Actuator and sensor electrodes are dimensioned in the range of 15-30 μm commensurate with experience in wired microelectrodes and as a compromise between switching cost (see above) and effectivity/sensitivity. A t-channel was chosen for the chemical solution partial containment geometry, to allow flexibility in electroosmotic and electrophoretic control of chemical concentrations based on prior experience with microfluidics [12], and to enable an investigation of potential future electro-locomotion using an electroosmotic drive [13]. With typical ionic and DNA oligomer diffusion coefficients of the order of $10^{-9}$ and $10^{-10}$ m$^2$s$^{-1}$ one can expect diffusional equilibration over 50 µm with a time constant of $\tau=L^2/D$ of 2.5 or 25s respectively. This is long enough for electrode induced species to show significant local accumulation. This partial isolation can be strengthened further by adding nozzles to the end of channel openings.

## 3. Lablet Fabrication (CMOS2)

The fabrication of lablets has seen successive simplifications and streamlining to reduce the number of processing steps, with a view to overcoming fabrication barriers, yield and cost effectivity. We show only the initial and final plans in what follows.

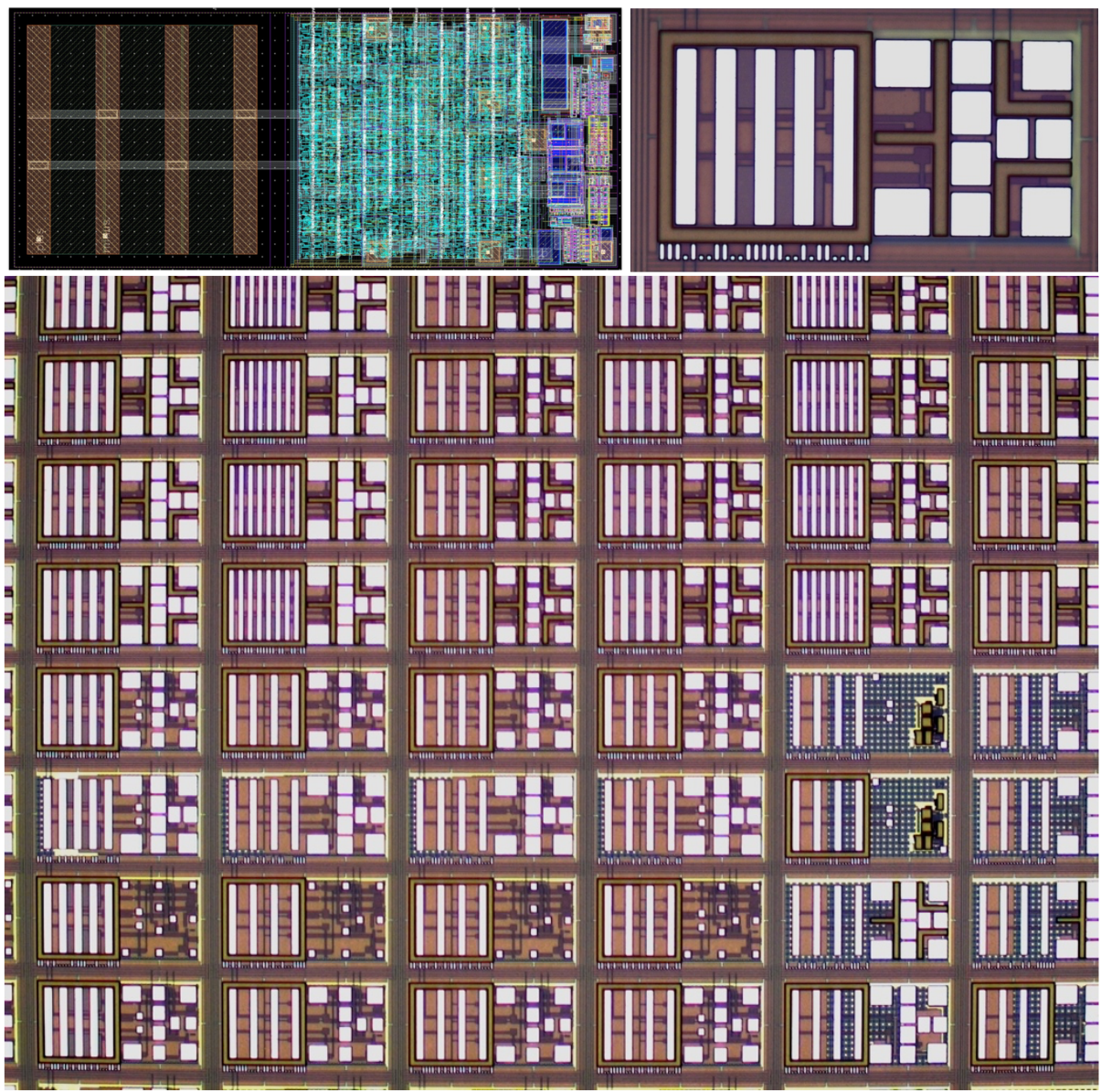

**Figure 3: CMOS2 lablet integration of physical and electronic layout. Top left:** With lablet dimensions of 200x100µm the physical design here shows the top metal in brown (interdigital structures for supercap clearly visible on left, actuator and sensor electrodes less so on right), the light blue digital electronic circuitry and the purple analog circuitry to interface with powering, actuators and sensors. **Top right:** The image shows open electrodes (white), raised walls for channels and reservoirs

(tan), galvanic network and other circuitry (bluish-purple). An optical barcode has been added to allow identification of the many combinatorial lablet variants. The lablets are separated by sawing along the galvanic bus lines (see processing step 2 below). The beginnings of supercap and channel walls are already present in the CMOS fab, using top metal underneath insulation to raise their foundations. **Bottom:** Array of lablets showing 4,5,8 fingered supercaps, different actor and sensor electrode sizes and analog test variants with minimum actuators (e.g. R5+6C5).

The first generation (named CMOS1) of such electronic lablets was designed to test lablet features without an integrated supercap and the lablets were fabricated on single chips of a multi-project wafer (5x5 mm, IMEC Europractice, 180nm CMOS technology TSMC). These chips would have required precisely aligned (and planarized remounting on wafers for efficient complex post-processing, without which only simpler non-photolithographic recoating of electrodes was performed. For example, M. Zoschke (Fraunhofer IZM) in a preliminary study found that lateral realignment of thinned 5x5 mm chips (35 µm) was achievable only to within 3-5 µm and reassembly was strongly limited by curvature in the chips arising from metal-silicon expansion tensions. Nonetheless, these lablets allowed a first version of the electronics for these microscopic particles to be tested, the height profile of the CMOS wafers to be ascertained and the establishment of a singulation strategy (using dice before grind wafer sawing/thinning) that formed the starting point for the strategy employed in later generations of lablets. Closeups of the CMOS2 lablets fabricated at TSMC are shown in **Fig. 3**.

## CMOS2 electronic design and fabrication

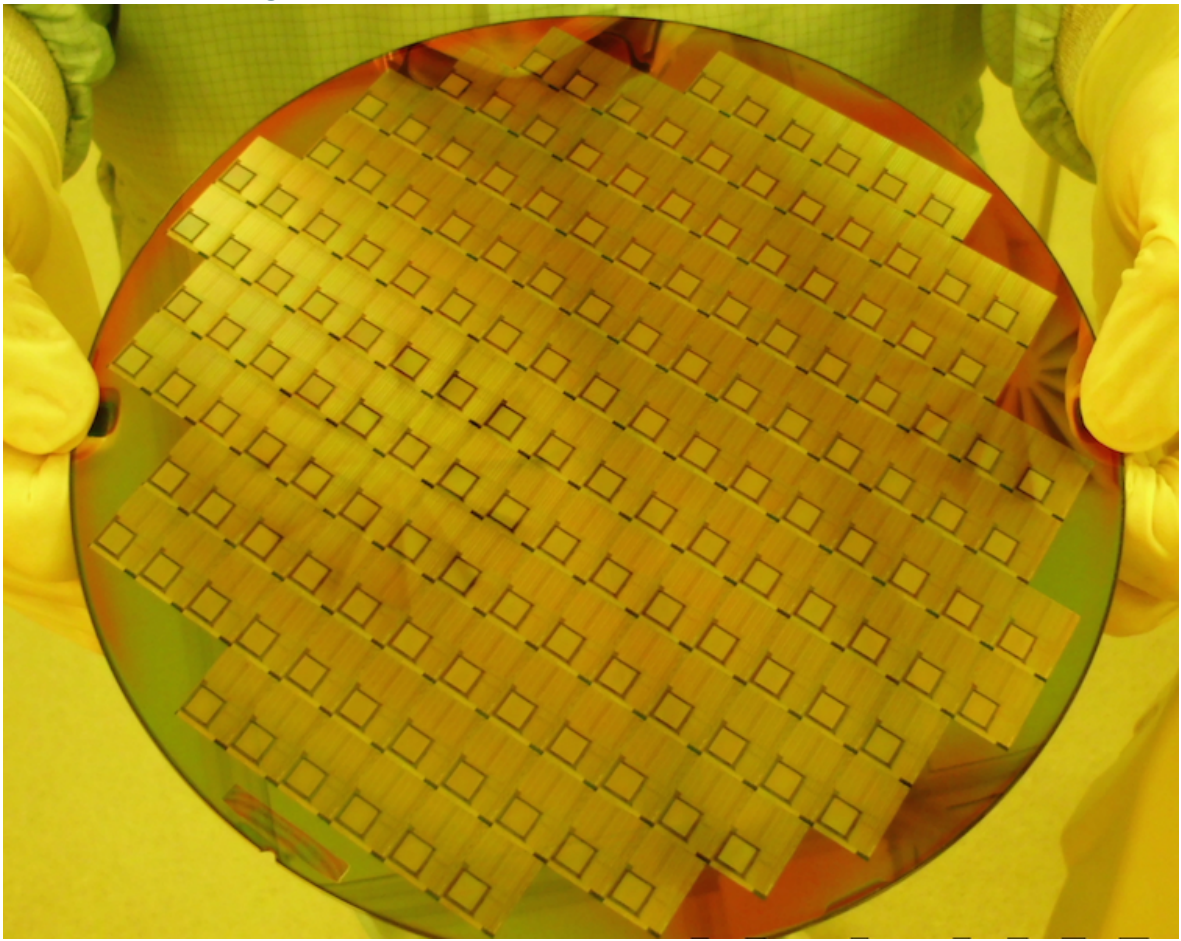

**Figure 4 8" wafer with lablets and docks**, designed by authors, fabricated by IMEC/TMSC in 180nm resolution CMOS. The wafer is fabricated using a stepper in 150 identical reticles, each containing several thousand lablets and one dock (dark squares).

For these reasons, CMOS2 lablet fabrication commenced with a technology run of whole Si wafer CMOS (engineering run, 12 8" wafers, 180 nm technology, IMEC Europractice TSMC), resulting in wafers as shown in **Fig. 4**. 8-inch (20 cm) diameter wafers were chosen as a compromise between standard CMOS fabrication efficiency (low end of size range) and post processing availability (high end of size range for prototyping facilities). The wafers were fabricated with 6 metal layers, and the thinner option (800nm) for top metal (Al) was chosen to reduce the magnitude of height variations in the surface. Note that there is no planarization after the top metal is structured, the top passivation (insulation) layer involving $SiO_2$ and SiON is applied uniformly and then the pads (i.e. all open metal structures, including supercap interdigital structures and microelectrodes) are opened locally through this layer. This results in three different surface heights: (i) metal surface (ii) passivation only (iii) passivation above top metal. Faced with this situation, other researchers requiring planar sensor surfaces have invested in an (expensive) initial planarization using CMP (chemical mechanical polishing) and construction of vias before further post-processing. Instead, we attempted to restrict top metal to only opened structures – supercap interdigital structures and microelectrodes, not using memcap structures for example – thereby reducing height variation issues primarily to the task of appropriate post-filling of recessed pads. In **Fig. 5** we show the measured height profile in the previous generation of chips (CMOS1), before this was taken into account. More information about earlier CMOS1 lablets can be found in [11].

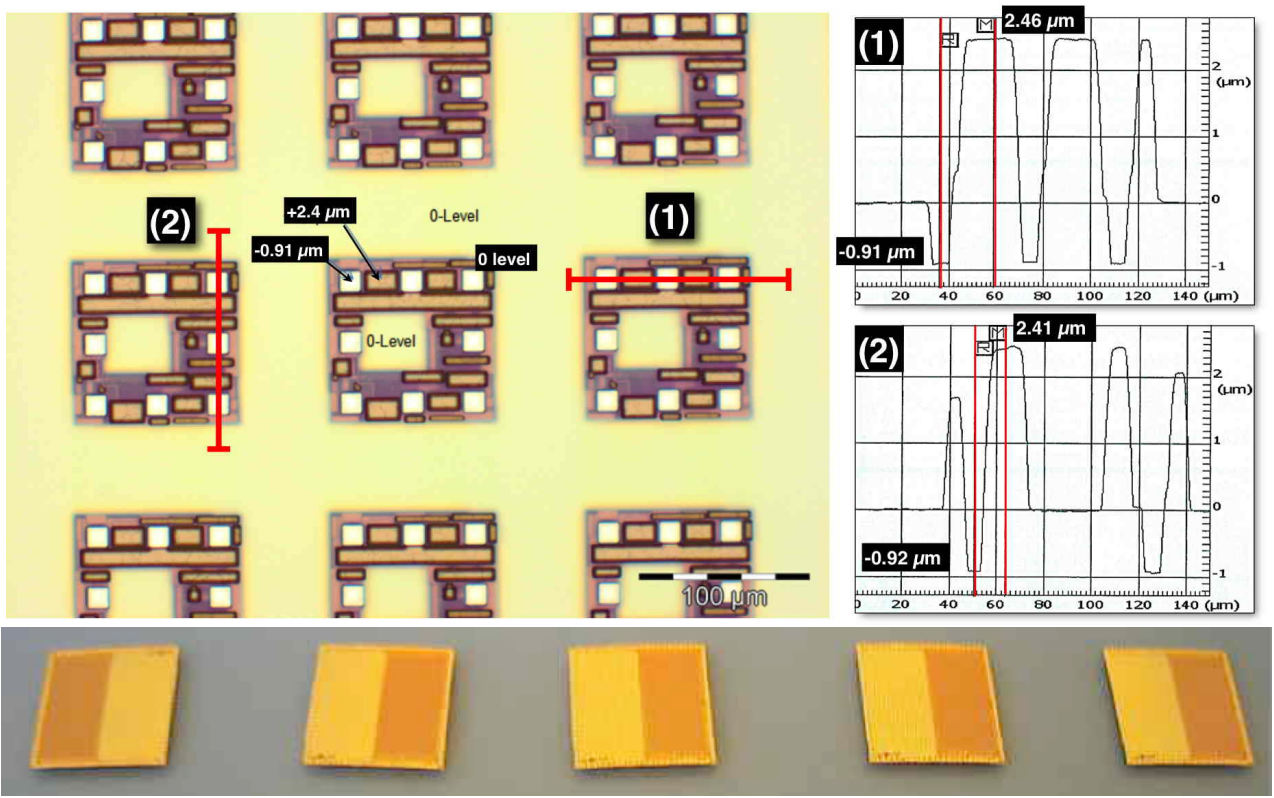

**Figure 5: CMOS1 lablet surface topology measurements.** The prior generation of lablets had filler metal and metal for memcap structures in top metal, resulting in large height variations of 2.4 µm. The measurements (1) and (2) of vertical height profiles along the cross-sections shown, were performed using a Dektak surface profiler (scan length: 150 µm). CMOS1 lablet microelectrodes, shown in white, are 12 µm square.

## Overview of post processing

The originally proposed post-processing scheme, as shown in **Fig. 6**, commenced with an integrated planarization and metal deposition procedure, and concentrated in intermediate steps on further planarization steps via matched metal deposition. It envisaged a solid-state (ionogel) electrolyte, enabling standard photoresist encapsulation and only used the dry photoresist laminate as additional protection and for suspended structures forming channels. Finally, singulation was to be achieved with a GBD (grind before dice) procedure.

This scheme emphasized the role of planarization (motivated by the large topography revealed in fig. 5) and was based on several assumptions including the compatibility of optimized supercap coatings with ionogels. The galvanic deposition necessary in step 5 required global metal interconnection of reticules in step 2, precluding a simple electroless coating of existing metal. Furthermore, the procedure would need significant extension to allow other microelectrode coatings (e.g. for sensors) at the same time. In addition, the proposed thinning before grinding procedure was found to induce significant tension (curvature) even at the level of 5x5 mm dies and would cause significant problems for whole wafers.

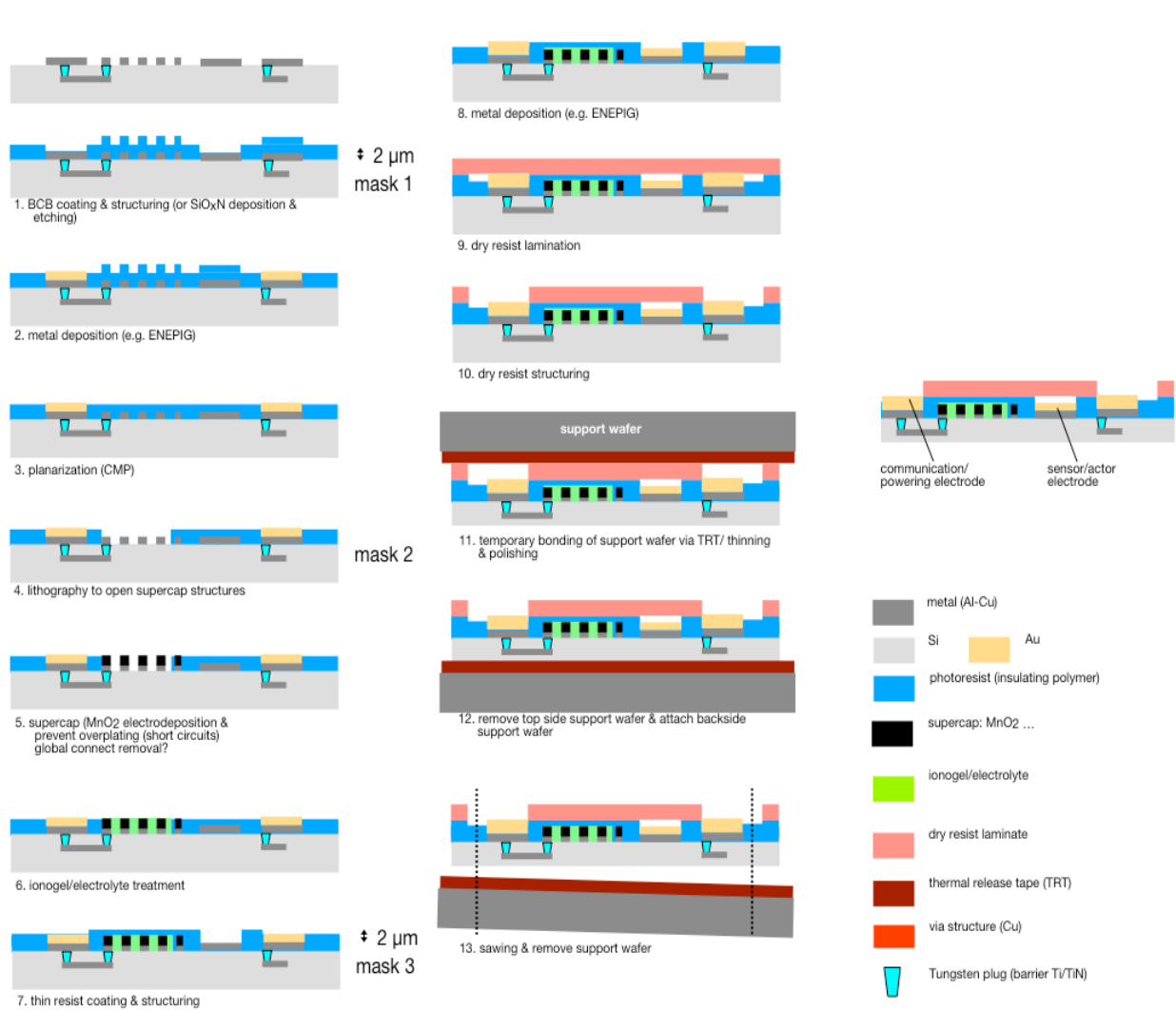

**Figure 6: Originally proposed post-processing scheme for lablets.** The process flow starts with planarization, metal deposition and reopening of supercap structures (1-4). Steps 5-7 coat and encapsulate a solid-state (ionogel) supercap. Steps 9-10 use dry resist photolamination to complete channels and supercap protection. Steps 11-13 implement one version of wafer thinning and dicing of lablets.

The final scheme arrived at is shown in **Fig. 7**. At the outset, it makes use of design elements in top metal in the the CMOS fabrication to minimize unwanted height differences and initiate height differences where needed at the base of walls. Metal balancing requirements in top metal are maintained while removing filler top metal from unwanted locations and pad sizes are carefully matched to pad openings to minimize insulation over metal protrusions. In step 0 it establishes approximate planarization via a simple maskless electroless metal deposition. In step 1, not required for single-reticule-based galvanic processing, global electrical interconnect of reticules to allow wafer-scale galvanic deposition was performed using PVD (physical vapor deposition) of Au/Cr and liftoff on a structured photoresist (mask M1). In step 2, a temporary protective (positive) photoresist was applied to this interconnect, and to metal structures not requiring galvanic coating, and then differential galvanic processing of the separate subnets of electrodes with identical functionality was performed. In step 3, the protective photoresist was removed and the wall structures for supercap and channels built using a permanent photoresist (SU8) or alternatively a photolaminate as in step 4. In step 4, electrolyte was added and a photolaminate (dry photoresist) used to seal it in to the supercap, completing also the roofs for lablet channels. In step 5 the lablets are thinned and singulated by sawing and/or laser dicing. Height profiles of lablets at step 3 are shown in **Fig. 8**.

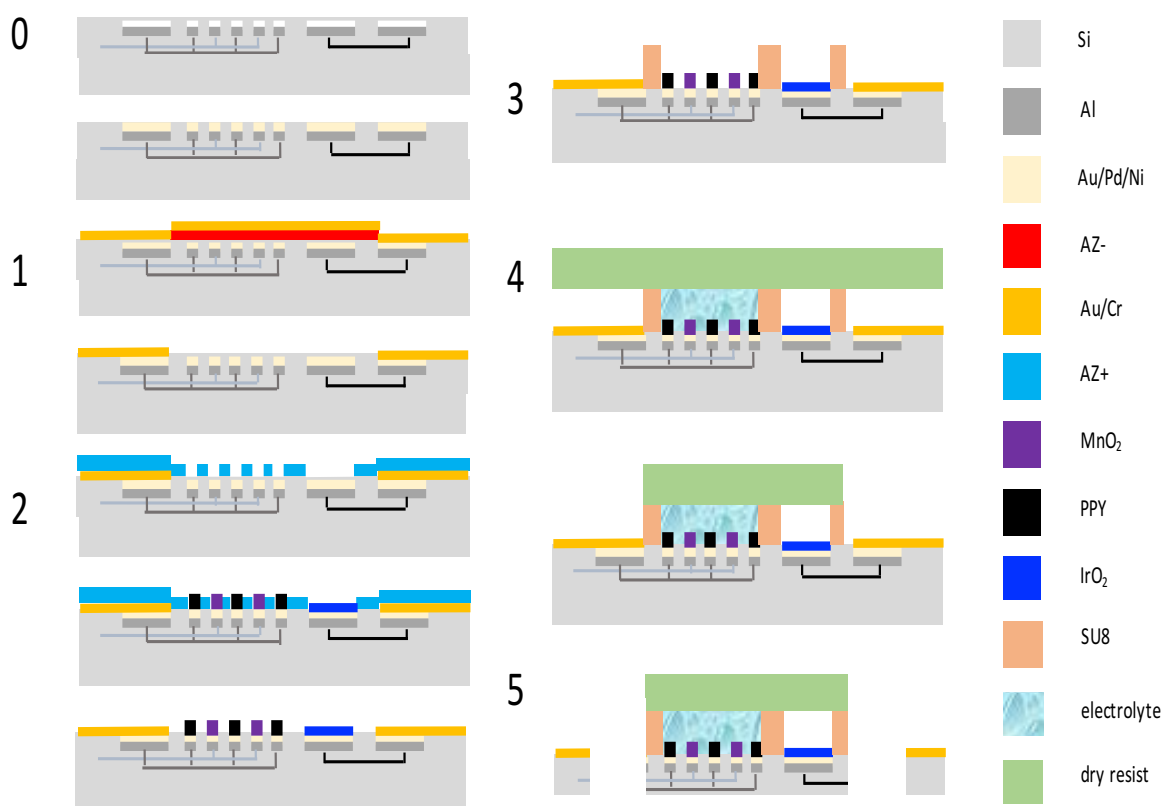

**Figure 7: Optimized post-processing scheme for lablets.** The process flow starts (0) with improved planarization in CMOS design and electroless metal fill deposition to gold coated microelectrodes, and proceeds with (1) global metal interconnect for galvanics (2) positive resist protected differential galvanic coating (3) SU8 walls above the CMOS elevated foundation (not shown) (4) electrolyte encapsulation and photostructured lamination (5) wafer thinning and sawing for singulation of lablets. Step 1 is not required with individual reticule galvanics: this is more controlled but less efficient (see text).

### Step 0: Electroless Gold Coating & Pad Fill
Since Al is also unsuitable for microelectrodes in aqueous solutions, the first step of post-processing is a remetallization and approximate planar filling of the Al pads. In order to preserve the initial sub μm resolution of structures, avoiding mask alignment errors, and for simplicity, we adopted a maskless ENEPIG procedure (electroless nickel, electroless palladium, immersion gold). The 800 nm recessed Al electrodes were coated with thicknesses of Ni 450 nm, Pd 250 nm und Au 70 nm, designed to nearly planarize the metal surface, for efficient electrochemistry and docking of lablets. The ENEPIG procedure was carried out by the specialist firm Atotech GmbH (M. Richter, Berlin) and resulted in a fine shiny gold finish with sub-micrometer granularity. Although the gold layer is quite thin, and susceptible to electrochemical dissolution, especially under strongly acidic conditions in conjunction with chloride ions, thicker gold layers become increasingly subject to adhesion limitations, peeling off from the underlying metal.

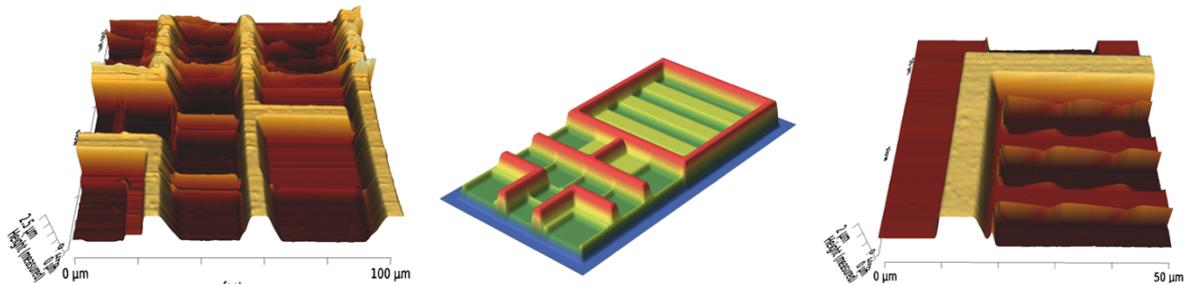

**Figure 8: AFM height profile of partly processed lablet. Left:** The height profile of an incompletely planarized lablet (pads filled with only 500nm Ni/PD/Au by ENEPIG) was measured using AFM. **Center:** calculated profile from masks. **Left:** AFM image of active area. **Right:** AFM image of part of the supercap. The steep high walls introducing some artefacts associated with rastering along fast axis: visible in the right image.

## Step 1: Reticule global interconnection for galvanic processing

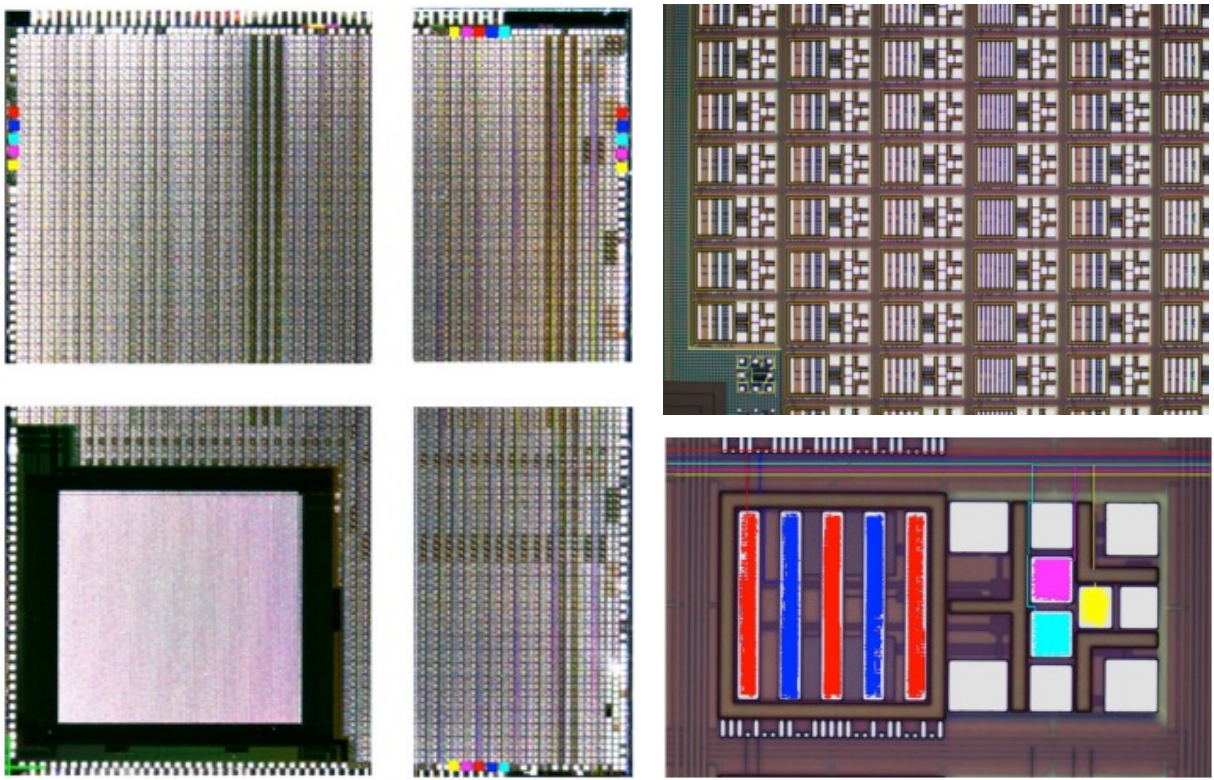

**Figure 9: Reticule with 5 galvanic nets. Left:** The reticule as cut into four for lablet array and dock extraction, showing the 5 colored galvanic network connections on each edge of the reticule. **Right:** increasing magnification showing (**top**) lablets with different geometries (supercap spacing) and (**bottom**) a single lablet with global "galvanic" interconnect (color labelled as for reticule interconnect

pads in left image). The networks interconnect in a mesh and use the different layers of CMOS metal to allow crossing without contact for the 5 independent networks.

The reticules were designed with 5 separate galvanic nets interconnecting all of the lablets except those not targeted for galvanic processing. The 5 nets, as shown in **Fig. 9**, connect respectively to the two poles of the interdigital supercap and to the three microelectrodes involved in lablet sensing: S1, S2 and Sref. Typically, for example, the two poles of the supercap can then both be coated with galvanic gold (to increase the surface area) and then separately by applying potential separately in successive solutions with $MnO_2$ and PPY (polypyrrole). This combination increases the voltage window of the supercap from < 1V for symmetric $MnO_2$ to ca. 1.4-1.5V for the asymmetric $MnO_2$/PPY system [14]. The sensor electrodes for pH detection for example can be chosen robustly to be $IrO_2$ with a Au reference [15-17]. The two sensors can be coated differently as required: any compatible galvanic coating can be employed. For example, enhanced polyoxometalate [18] or NP coatings [19] may also be deposited galvanically.

In order for all reticules to be coated at once efficiently, these five nets need to be connected globally to the periphery of the wafer (see galvanic processing in step 2). A PVD lift off procedure with 200 nm Au with Cr as an adhesive intermediate was deemed to provide the best chance of reliable connections between the 100μm square pads at the periphery of neighbouring reticules. This step was carried out by iX-Factory (Dortmund, H. Bouwes) using a double layer photoresist LOR/PMGI (MicroChem), based on based on polydimethylglutarimide.

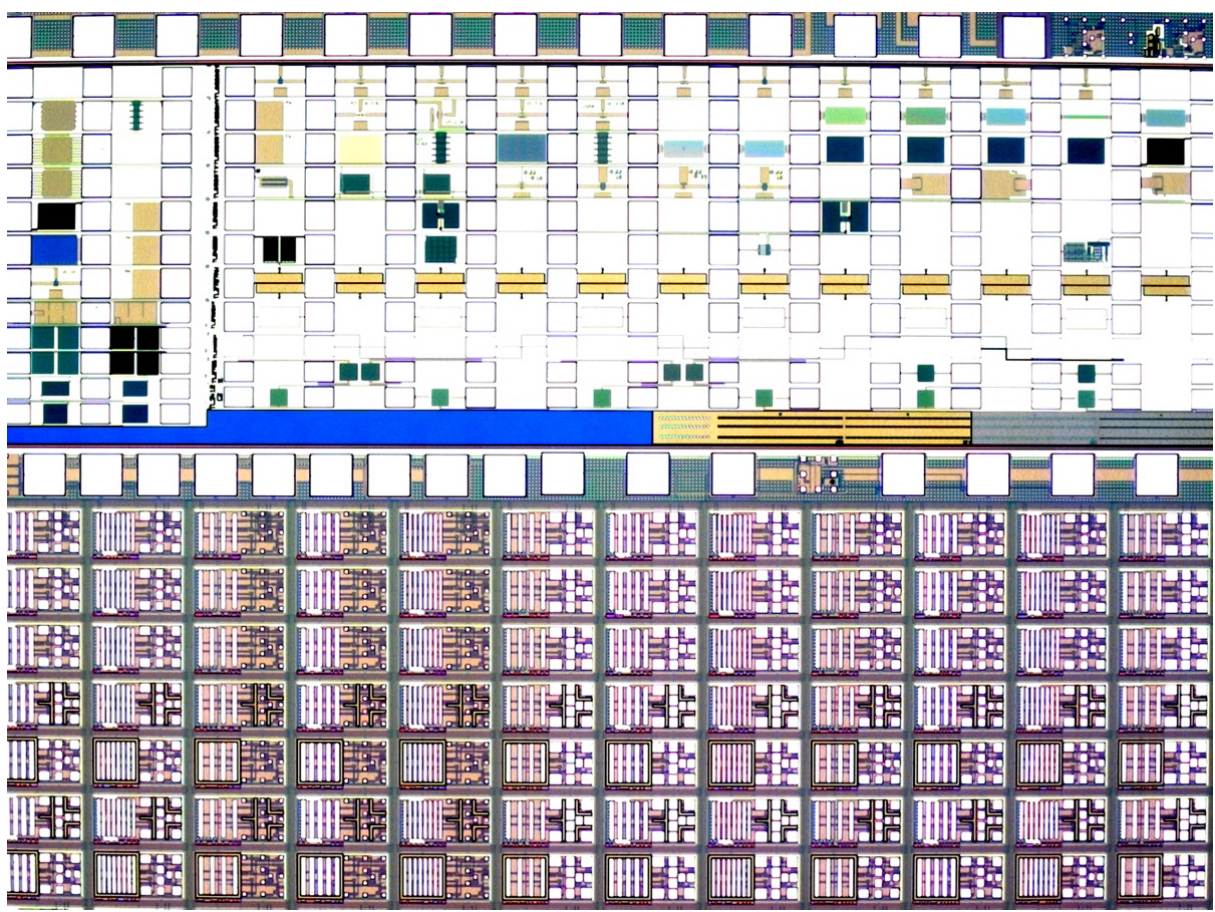

**Figure 10: Path for inter-reticule connections.** The image shows the top edge of one reticule, with a row of 100 μm square pads (white), 5 of which need to be connected to opposing pads at the base of the next reticule (top of picture). In between, the space is filled with test structures, showing a topography as in fig. 5.

A mask (M1) was designed with 5 μm minimal feature size using scripted automation software and manufactured as a 9" chrome mask on borosilicate glass (Photronics GmbH). The wafer fabrication process (TSMC, Taiwan) requires extensive test structures in the space between reticules (see fig. 10), which needs to be crossed by the inter-reticule connections. These structures have topography with residual height differences (after ENEPIG) of 1.5 μm or more and steep edges. Initial tests on interconnecting test wafers (top metal only from TSMC) revealed unreliable connections, with only *ca.* 30% of reticules connected. For this reason, the gold thickness was increased to 275 ±25 nm (higher values result in peeling) and a twin-angle PVD process was agreed with iX-Factory. The resulting interconnection traces, routed between the TSMC test structures (**Fig. 10**) appeared optically continuous, but showed several dark bands where the trace crosses sharp elevation differences in the substrate.

The structures were then measured for electrical resistance using a probe station. The results confirmed that connections were still not reliable, especially near the edge of the wafer. For example, the reticules nr. 57, 58 and 59 at the extreme right of the 5$^{th}$ row from the top (cf. Fig. 1) showed no horizontal connections between 2 of the 5 traces between 58 and 59, and resistances ranging between 700 and 7400 Ω between 57 and 58. The resistance could be correlated with the number of steps in intervening test structure topography that the traces had to cross, confirming the difficulty in metalizing the steep walls of these structures via PVD. A check of resistance across partial traces confirmed this conclusion. By contrast, measurement of resistances between corresponding pads on opposite edges of the same reticule revealed values of 27-37 Ω, consistent with probe resistance only, i.e. confirming full conductivity of the TSMC fabricated multi-level traces. Similarly, the vertical connections which are twice as long (*ca.* 1 mm) were investigated between the outermost reticule 59 and 44 (above it). Three of the five connections were not connected, one at 7 kΩ and one at 43 Ω. While the situation in the centre of the wafer was significantly better, the overall performance was insufficient to allow reliable connection to the majority of reticules, despite the redundancy built into the rectangular mesh. Thermal expansion cannot account for these issues. Differential wafer vs. mask thermal expansion and wafer tension following processing lead to mask alignment errors of at most 2.5 μm in the outer regions of the wafer, and these are insufficient to account for the problems in connectivity. Non uniform deposition probably contributes, but tests of connectivity between large contact areas for the galvanic bath at the periphery of the wafers revealed good metal contact (30 Ω).

Alternative metallization and connection schemes were considered (D. Bouwes, iX-Factory), including wet chemical etching of global metal, but preliminary tests at iX-Factory (A. Kasjanow) revealed that the etching process caused severe damage to the underlying metallization and electronics. This step could be resolved in future through the inclusion of an additional planarization process, either by the wafer foundry or in separate post-processing. The full extent of the problem only became apparent by degrees, and it could not be resolved on the timescale and with the budget available to the project. As we will see in the next section, an alternative reticule-based galvanic strategy was then developed and pursued.

2: Protection and programmable galvanic coatings
The thick (several micrometer) galvanic coating ($MnO_2$ and PPY) of supercap interdigital structures with comparatively narrow gaps, down to 5 μm, poses a serious risk of short circuiting connections if galvanics is pursued directly. For this reason, the interdigital structure is reinforced with a reversible photoresist, building a 4 μm micrometer high protective wall between adjacent electrodes. Such protection is also valuable in creating barriers between closely spaced sensor electrodes that need to be coated (for example with $IrO_2$). This poses strict conditions on mask alignment, since misalignment will detract from the available area contacting the metal interdigital structures. A positive resist (AZ+) was employed and the photolithography carried out by Temicon GmbH (Dortmund). Mask alignment errors were reduced to ≤ 2 μm for 90% of the wafer, and ≤1 μm in the central region (see **Fig. 11**).

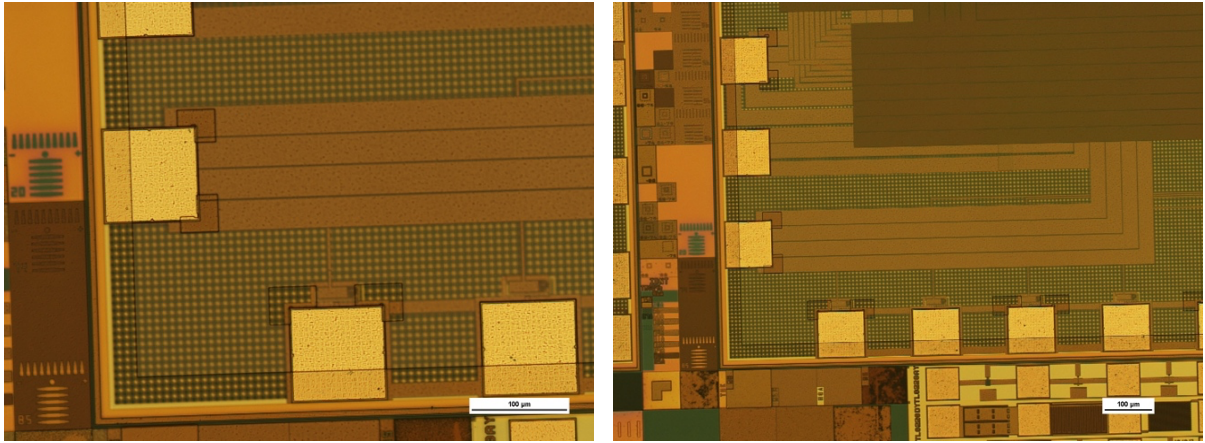

**Figure 11: Alignment of protective mask (M2) to metal structures.** The two images show the alignment quality at different points on the wafer. The misalignment of the mask is primarily because of expansion coefficient mismatch between the processed wafer and the borosilicate chrome masks, despite computed matching expansion factors for silicon.

Although galvanic coating is partially specific to electrodes provided with a potential, it is also advisable to cover structures that should not be galvanically coated, especially if prolonged or multiple coating is required. The mask 2 was also used to insulate both actuator, power and communication pads and the global interconnection structures.

A customized galvanically connected wafer holder for 8" wafers with a multinet connection scheme designed by the authors was purchased from Silicet GmbH. The wafer holder consists of a flat base and an upper ring made of polypropylene positioned by two pins and held together by plastic clips, see **Fig. 12**. The base and ring hold an 8" wafer, aligned at the notch, in a sandwich (covering the outer 5 mm of the wafer) and achieve water tight connection at the inner lip of the ring using a specially contoured inset gasket (EPDM rubber). Electrical connections are made in the dry zone by means of thin metal plates connecting to the outer portion of the wafer at about 3mm from the periphery. A special wiring scheme, connecting these plates separately to 8 external wires, allows in future up to 8 subnets to be independently activated for galvanic processing. The position of the metal plates can be readjusted to match the position of global nets on different wafers.

Initially conceived to allow combinatorial tests with different coatings on different reticules, while preserving some level of efficiency, the translatable adaptor module proved essential to allow systematic galvanic coatings to proceed in the absence of reliable global interconnect (see step 1). The adaptor consists of 11 spring loaded prober needles (Bürklin GmbH, Oberhaching, Germany, F109.18) allowing up to 6 reticules to be contacted in parallel. The prober needles are mounted in a precision-drilled PMMA block, with two needles per reticule (for 5 of the six reticules) to allow conductive check of correct contact. Only one set of reticule connections (vertical) were used, to allow translation of the module to shift between different galvanic nets in parallel for all 6 reticules. The adaptor was mounted on a motorized z-axis, equipped with a mechanical screw mount to allow fine rotational adjustment of the needles to the orientation of the wafer. Beneath the PMMA block (1.5-2 mm) a counter electrode (Ti mesh, optionally coated with Pt) is suspended using Teflon screws and spacer rings. Additionally, a Ag/AgCl reference electrode is suspended from the side of the adaptor module.

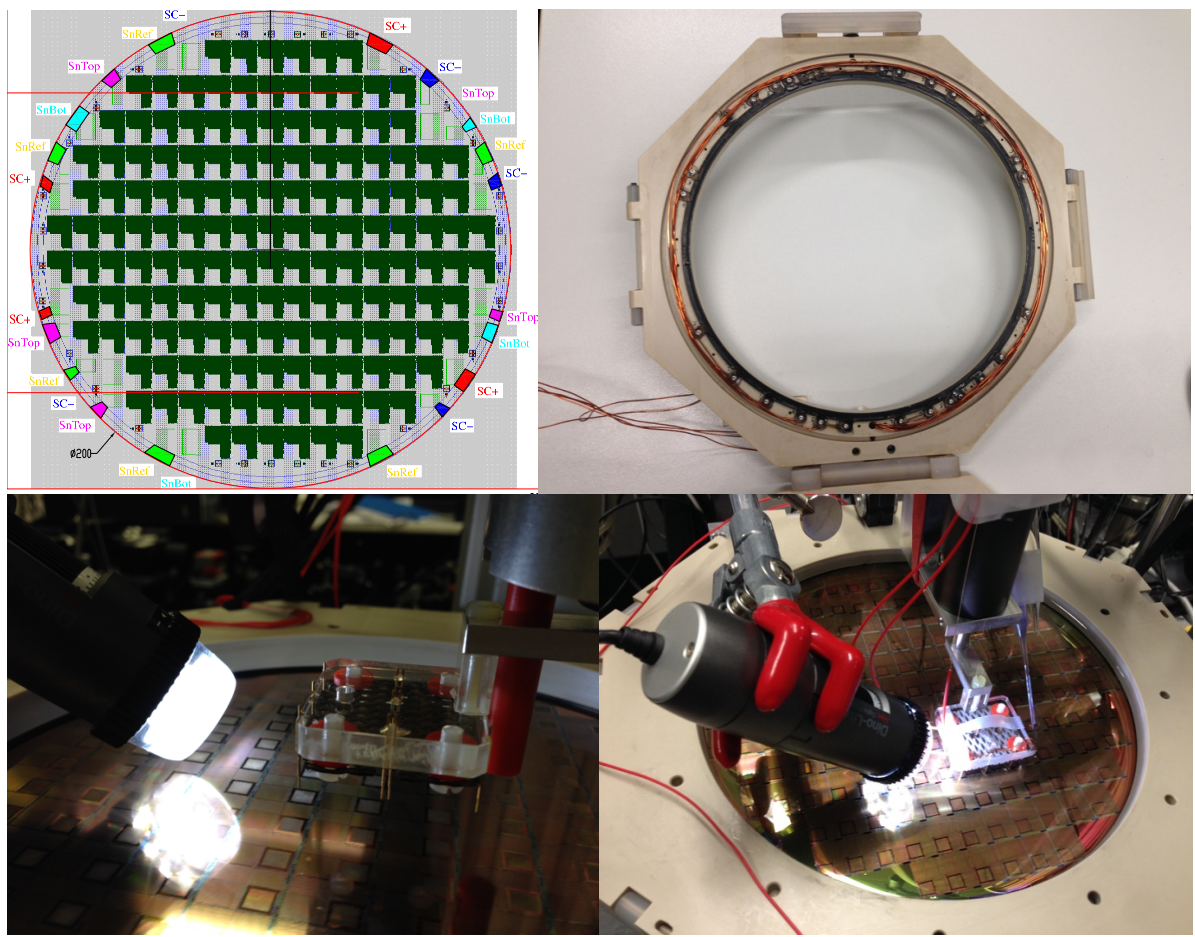

**Figure 12: Galvanic processing on wafer and reticule scale**. **Top left:** reticule interconnection structure for five galvanic nets, each connected redundantly to four conducting surfaces at the periphery of the wafer, allowing independent coating of halves of the supercap and sensor and reference electrodes. **Top right**: 8" wafer holder (Silicet GmbH) adapted for multiple galvanic net wiring. **Bottom left and right**: images of multi spring-loaded prober needle adapter to contact one galvanic net on up to 6 reticules, attached to a motorized z-axis with the wafer holder mounted on a 200mm travel xy-stage with 1μm resolution. Also shown is a USB microscope for position monitoring. A Pt-coated Ti mesh is used as an "anode" for galvanic coating of metals (cathode for $MnO_2$ and PPY deposition).

The protruding top side of the prober needles are soldered to wires and connected alternately to a multimeter for contact (conductivity) monitoring, or to a multifunctional potentiostat (Autolab III) for current or potential controlled coatings. The absence of insulation on the prober needles meant that they too were coated during galvanic processing and needed to be cleaned between batch runs. The module was lowered and raised vertically to achieve electrical contact, following alignment with the wafer. After completion of one set of reticules, the module was translated to neighbouring 6 reticule blocks for 10 to 15 times to coat up to 2/3 of the wafer. The galvanic solution is then exchanged with a coating solution for one of the other galvanic nets, and the process repeated. An example of galvanically processed lablets is shown in **Fig. 13**.

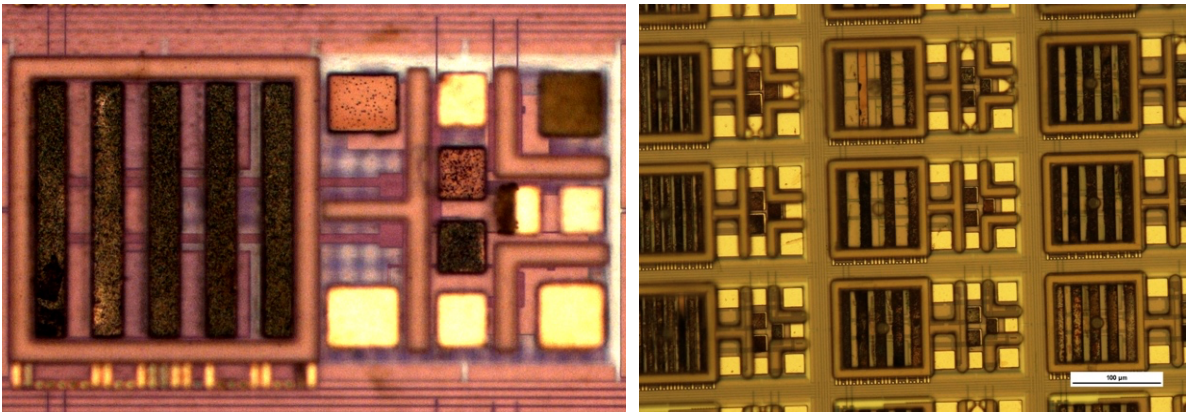

**Figure 13: Galvanically coated lablets on wafer with SU8 walls. Left:** Single lablet after galvanic coating with $MnO_2$. **Right:** Nine lablets can be seen with $MnO_2$ coated supercaps (step 2) and partially coated sensor electrodes. The SU8 wall structures (step 3), including nozzle structures in the first row can also be seen, well aligned with the top metal foundation.

This process allowed the three different coats of $MnO_2$, PPY and $IrO_2$ in addition to the already present Au to be deposited. Solutions and electrochemical procedures are documented in the SI. Details of the optimization of the coatings for supercap functionality can be found in [8,9].

Unfortunately, in the processing of the CMOS2 wafers, the protective photoresist tended to release debris on needle contact which tended to insulate the connections making the process of establishing contact very unreliable. For this reason, the protective coating had to be removed from several of the initial wafers to allow reticule-level coating to proceed. The AZ+ photoresist can be removed effectively and rapidly by rinsing in acetone.

### 3: Permanent photoresist construction of supercap and channel walls

SU8 photolithography (mask M3) was carried out (Temicon GmbH, Dortmund) to deposit 7 μm high SU8 walls above the CMOS wall foundations (elevated 1μm by insulator underlying top metal). The mask for this process was designed to allow the walls to function within the tolerances of the measured alignment errors (in the range of -2 to-2 μm). The base of the walls was typically 9 μm wide giving the walls an aspect ratio of 0.78. Since the adhesion of SU8 to the substrate is critical both for effective lamination (step 4) and stable dicing (step 5), attempts were made to improve the stability of adhesion by (i) using short plasma activation in $O_2$ and a two-step SU lithography (thin, thick). Proper cleaning of the substrate, including complete AZ+ removal prior to spin coating is essential. As an alternative, dry photoresist lamination was also performed, with a 10 μm thick photolaminate (Nagase, 1010), but although high aspect ratio structures can be achieved, these walls were not as well defined as for the SU8 structures. An example of the well-aligned SU8 wall structures is shown in **Fig. 14**.

The first mask used for this step only had SU8 for the walls themselves, and hence had a very low percentage fill of the area. This proved problematic for the lamination process (see step 4) where the adhesion of these structures has to be sufficient to compete with the full area adhesion of the laminate liner during peel off. For this reason, a modified second mask (M3b) was employed that introduced additional walled structures between lablets and over the docking chip (also to be used as test structures on the dock [20]).

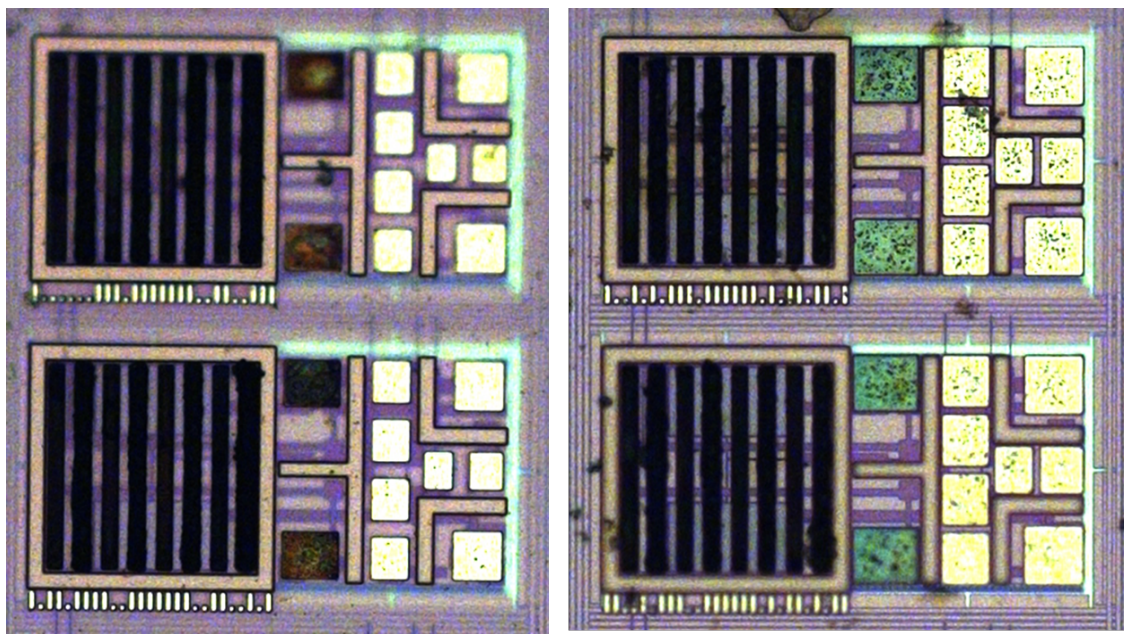

**Figure 14: Galvanically coated lablets on wafer with SU8 walls and additional filler structures.** Compared with fig. 13, the additional structures are designed to improve adhesion of laminate during peel off (step 4), see text.

## 4: Electrolyte encapsulation and photolamination of channel roofs

Electrolyte encapsulation is necessary to separate the high ionic strength solution, required to deliver high capacitance, from the more dilute aqueous solutions that the finished lablets should operate in. Although solid state electrolyte systems, e.g. ionogels using PVA are attractive for acidic supercaps, they proved less compatible with photolamination (also needed to create channel structures) and with the $MnO_2$ based supercaps developed in this project [8], which cannot operate in strong acid. Instead, we employed a 50% glycerin solution with a small quantity of surfactant (0.25% triton) in $NaSO_4$ (symmetric MnO2 supercap) or $KNO_3$ (asymmetric MNO2/PPY supercap). The glycerin at high concentration makes the aqueous solution hygroscopic, limiting evaporation and ensuring sufficient processing time and shelf life of the lablets. Its effect via diffusion on limiting capacitance is about a factor of two in macroscopic tests.

The choice of photolaminate is in part dictated by availability, since prototyping requires comparatively small quantities. Earlier work on passive test lablets with the photo laminates – TMMF (S2020, Tokyo Ohka Kogyo, Japan) [21] and ordyl (FP415, Elga Europe, Italy) which is an alternative product to PermX, (DuPont, 14-20 µm) [22] – demonstrated that aqueous solution could be encapsulated in recessed structures during lamination on nickel coated substrates. Whereas the Dupont photolaminate subsequently became unavailable in low thicknesses (<15 µm), another high aspect-ratio dry photoresist family, the Nagase DF1000 (or 2000-4000) series (Engineered Materials Systems, Inc., USA), was obtainable in thicknesses down to 10µm. The Nagase dry resists are negative photoresist with a low adhesion temperature of 45° C and moderate soft and hard bake temperatures (100°C and 175°C) which can be lowered significantly (here to 80°C) at the cost of longer bake times. Illumination fluxes are moderate (especially for the thinner films (down to 100mJ/cm$^2$) and development is straightforward in less than 5 minutes in cyclohexanone. While the higher series 3000 and 4000 have increasing hydrophobicity, this can hinder the distribution of aqueous electrolyte during lamination. Most lamination experiments were performed with DF1010 (10µm), with the more hydrophobic DF3014 (14µm) also tested.

A large area heated roller laminator (Fetzel, Austria, SDL 50) was employed in the cleanrooms of Temicon GmbH (Dortmund) for wafer lamination. For lamination tests, a row of electrolyte droplets with varying composition for tests, sufficient to wet 2/3 of the wafer during lamination, was placed across the 8" wafer (parallel to the roller) leaving the first third of the wafer free of electrolyte in order to compare with electrolyte-free lamination. The laminator rollers were heated to 80°C, the gap adjusted to 1.5 mm (somewhat less than the 740 µm wafer plus 1mm compressible plastic sheet carrier combined thickness) and a slow lamination speed selected 0.3-0.4 m/min. After removal of the first protective liner from a 25x25 cm square of laminate cut from the roll, lamination proceeded without the table, keeping the laminate raised with the remaining liner against the first roller during wafer entry. Larger laminator gaps or lower temperatures gave insufficient adhesion during liner peel-off and higher temperatures caused excessive evaporation of the electrolyte. Initial tests were carried out on non-CMOS test wafers, coated with the same SU8 wall structures to mimic the recesses of the true lablets. These test wafers were cut into quarters for separate lamination. In addition, unstructured 6" silicon wafers were employed in first tests with the new laminates. Wafers were passed twice through the laminator, and then excess laminate was trimmed with a scalpel. The wafers were then heated to 60° for ca. 1 min and allowed to cool before peeling off the remaining liner (at a 90° angle).

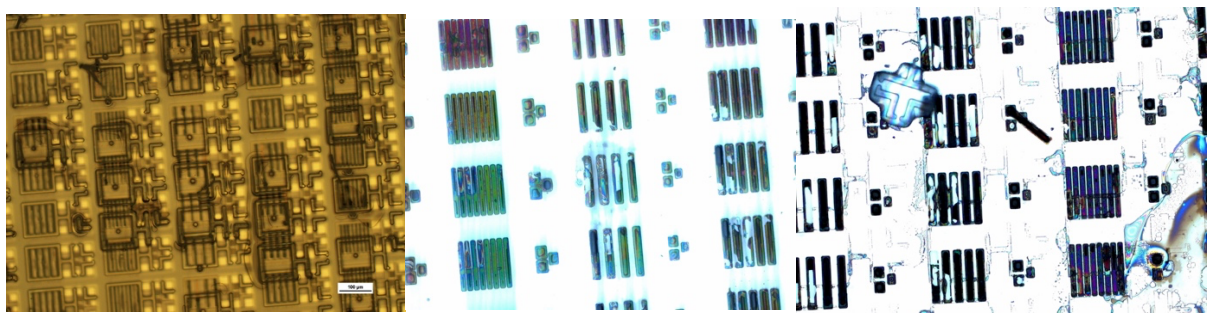

**Figure 15: Peel off of structures with liner removal.** Left: Peel up and reattachment before photolithography results in shifted structures. Middle: lifted galvanic coatings to liner. Right: removal of SU8 structures and galvanic coatings. Problems overcome with revised mask 3b (see text).

As referred to previously in step 3, the first mask used for the SU8 walls was insufficient to prevent lifting of the laminate during liner peel off in the presence of electrolyte. Adhesion could be weakened by the wetting of the top of the SU8 walls during electrolyte dispersion under the rollers (the quantity of surfactant in the electrolyte for wetting of the enclosures was reduced as far as possible, see above) and by the lamination pressure on the enclosed supercap container (which would argue for lower lamination pressures, i.e. larger gaps). Attempts to improve adhesion by peeling off at elevated temperatures (80°C) failed. Examination of the peeled off liner confirmed that laminate lifted primarily in electrolyte wetted areas and that in most cases the SU8 wall structure and in some cases the $MnO_2$ supercap coats were also lifted off (see **Fig. 15**). The latter is a sign of over compression in the supercap enclosure, since these layers should be several micrometers in fluid below the laminate. The former (SU8 liftoff) was initially surprising, given good experience with SU8 adhesion in earlier microfluidic experimentation [23], but understandable in retrospect because of the extreme differences in surface area of the SU8 walls in comparison with standard microfluidics (here only 1-3 % of the total area). Additional SU8 structures were introduced in step 3 to increase the coverage fraction to >10%, thereby decreasing forces on the SU8-wafer bonding by a factor of 5 or more. This showed a significant improvement, with a significant reduction in SU8 rip-off and in lift and settle translations of the SU8 wall structure prior to illumination. The DF1010 liner must be removed prior to exposure.

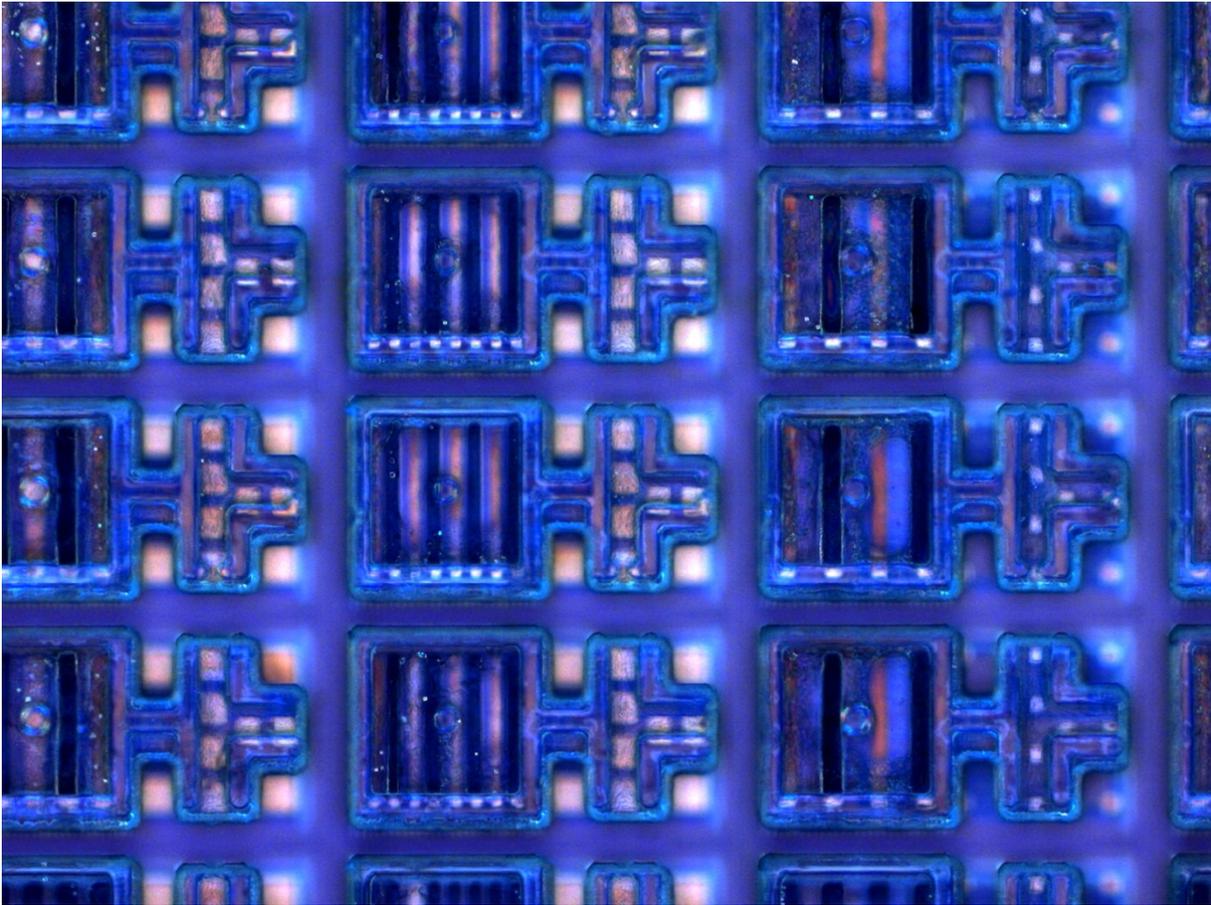

**Figure 16: Lablet structures following laminate development.** Four images showing laminated lablet structures: (i) without filler or electrolyte (ii) with electrolyte (iii) with filler and no electrolyte (iv) with filler and electrolyte.

Illumination was reduced from the 200mJ/cm$^2$ @ 365nm UV suitable for thicker films, to 100mJ/cm$^2$ for the DF1010 and led to well defined structures with sharp edges in the absence of electrolyte. The electrolyte appeared to cause some distortions in illumination, although it remains unclear whether this is primarily related to adhesion weakening, optical effects during photolithography or influences during development.

Following exposure, a soft bake at 80°C was performed on a hot plate for 20 min. The wafers were then developed using a soak and spray protocol for a total of 5 minutes in cyclohexanone in reduced light under a fume hood. The wafers were then rinsed with isopropyl alcohol and water and blown dry gently with clean compressed nitrogen. In lieu of a full hard bake at 175° for 1 h (optional to increase stability), an extended bake at 80°C for 1-2 h was performed. Note that higher temperatures than 100°C proved detrimental to $MnO_2$ capacitance and the electrolyte evaporates extensively at temperatures above 80°C. Resulting structures are shown in **Fig. 16**.

## 5: Lablet singulation

Lablet singulation was initially envisaged using reactive ion etching, but the conditions for this proved incompatible with soft lablet parts including laminate and electrolyte. Instead, a professional dice and grind service was employed (Disco GmbH, Kirchheim). First tests with the singulation of 5x5 mm chips with CMOS1 revealed that a dice before grind procedure was adequate to singulate lablets using a 20 µm saw in the 30 µm wide free zone between lablets. Although some chipping of the lablets occurred (see **Fig. 17**) the chip-scale singulation of these mostly unprocessed lablets (only step 0 completed, ENEPIG coating of electrodes) allowed full electronic functionality to be preserved, as checked by comparison with simulated electronic performance and by comparison with performance in lablet arrays [10].

The move to whole wafer lablet arrays and complex processing presented several challenges to the singulation. Firstly, a thin before sawing approach became impractical because of large area wafer stress and a dice before grind strategy was mandatory. The galvanic networks placed in CMOS metal layers in the saw lanes between lablets complicated the conditions for mechanical sawing, especially with finer blades, required for reduced chipping, being deflected by the metal lines causing blade damage. For this reason, Disco employed an initial laser process to cut through the top layers before continuing with mechanical sawing.  The second major challenge is adhesion of lablets to the carrier tape during thinning. The processed lablets must be attached via the uppermost layer, the laminate, providing a further adhesion requirement to those already discussed in step 4. For this reason, yields of lablets had to be increased by sensitive thinning protocols and increased from ca. 10% to over 95% for unlaminated lablets and over 60% for laminated lablets.

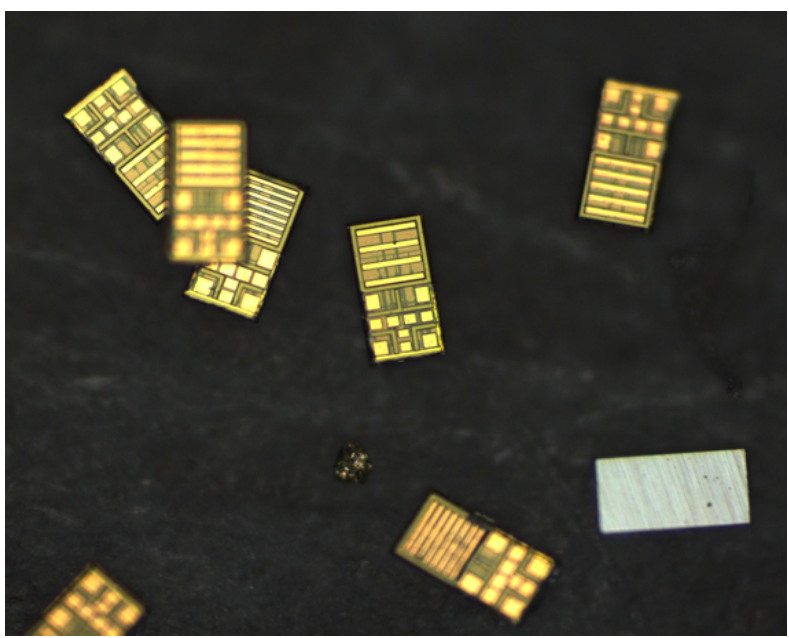

**Figure 17: Singulated lablets.** Initial lablet singulation was performed with the ENEPIG processed lablets. The process may be compared with singulation of the fully processed lablets in Fig. 22.

The results of lablet singulation are shown in fig. 17. Further improvement by shifting to an all laser cutting process are envisaged (see CMOS3).

## 4. CMOS2 Lablet Variants & Characterization

Lablet fabrication at the wafer scale involves identical reticules, but since each reticule contains several thousand lablets there is ample scope for combinatorial variation at this early stage of prototype development. Some of the major variants implemented already in the CMOS fabrication are shown in **Fig. 18**. These include both electronic variants and physical variants. The optical bar code (at the base of the supercap) encodes not only these variants but also a lablet ID which can be useful in controlled studies of lablet sorting for example.

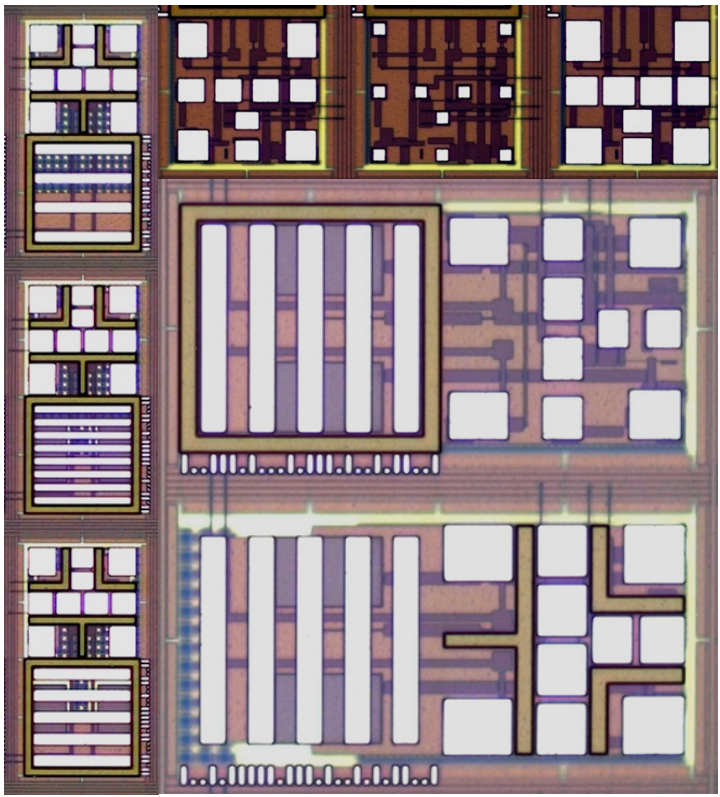

**Lablet variants**
- 6 analog lablets (6)
- 2*12 digital lablets (24)
- photodiode as data in (2)
- supercap structures (3)
- pad size (3)
- SC seal ring (2)
- channel wall (2)
- galvanic connect to supercap (2)
- galvanic connect to sensorpads (2)
- optical barcodes (above plus ID)

**Figure 18: Lablet variants .** Three examples of lablet variation are integrated in the image: left block, 3 variations in supercap interdigital structure, top 3 variations in analog lablets and electrode size, bottom right: with and without channel walls (as well as electrode size variation). Note that the lablets are the same size but scaled differently in the three examples. Note that the major option of channel walls here is at the foundation level and can be taken separately for the step 3 SU8 masked walls.

In addition to these variants, lablet post-processing can be done in different ways to yield further lablet variants. In particular, variation can be introduced at each of the four mask stages and in the galvanic processing and encapsulation of electrolyte. Examples of the former include options with and without channel walls and supercap electrode widths. Examples of the latter include symmetric vs asymmetric supercap coating (e.g. $MnO_2$ vs $MnO_2$/PPY, see step 2), variations in sensor coatings, variations in the presence and positions of channel walls, variations in the choice of electrolyte (can be done differently at different locations on the wafer during lamination) and variations in laminate geometry. Combinatorial optimization of coatings using the docking chip as a galvanic engine are being explored in a separate paper [23] but can also find application here.

Optical characterization of the lablets with SU8 walls (**Fig. 19**) and photostructured laminates (**Fig. 20** in closeup and **Fig. 21** on a larger portion of the array) are followed by optical characterization of the singulated fully processed lablets in **Fig. 22**.

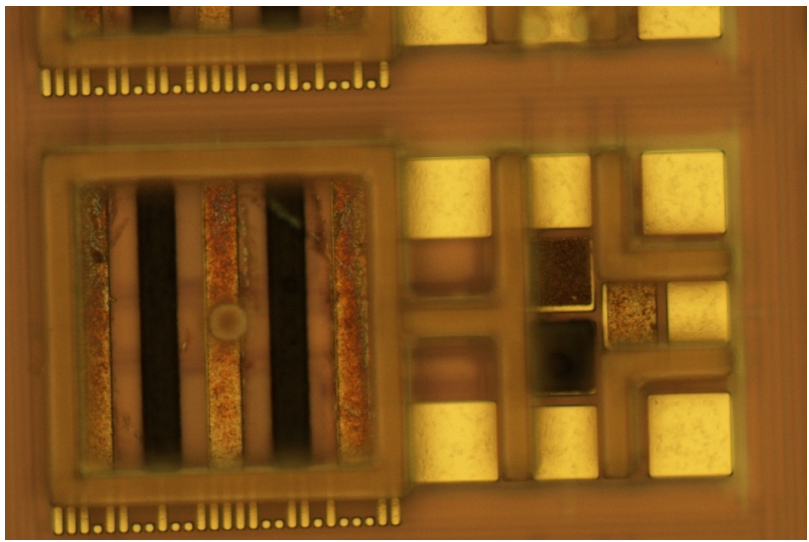

**Figure 19: Detail of lablet processed with asymmetric supercap galvanic coatings and SU8 walls.** This example shows the good galvanic discrimination with uncoated gold actuator and power electrodes, and the high aspect ratio SU8 walls well aligned with the foundation. One can also see the central round pillar in the middle of the supercap to prevent lid sagging after lamination. The reproducibility can be seen in the beginning of a second lablet at top.

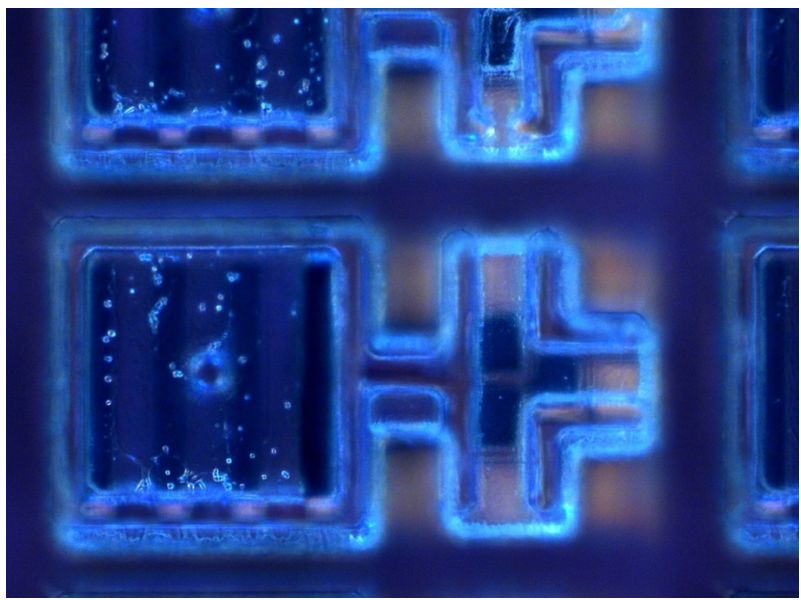

**Figure 20 Fully processed lablet with encapsulated supercap.** This example shows a laminated lablet on the wafer after photolithography and development removes the laminate from surrounding area, leaving an aligned seal for the supercap and a roof for the T-channel on the active part of the lablet. The coated supercap structures can be seen through the transparent laminate as well sensor and actuator electrodes on the right inside the channel. The 50x objective employed means that the plane of electrodes (17µm below the surface is out of focus). The beginning of the next lablet in the array can be seen. Alternate lablets have nozzle structures at the end of channels and these are visible in the upper lablet.

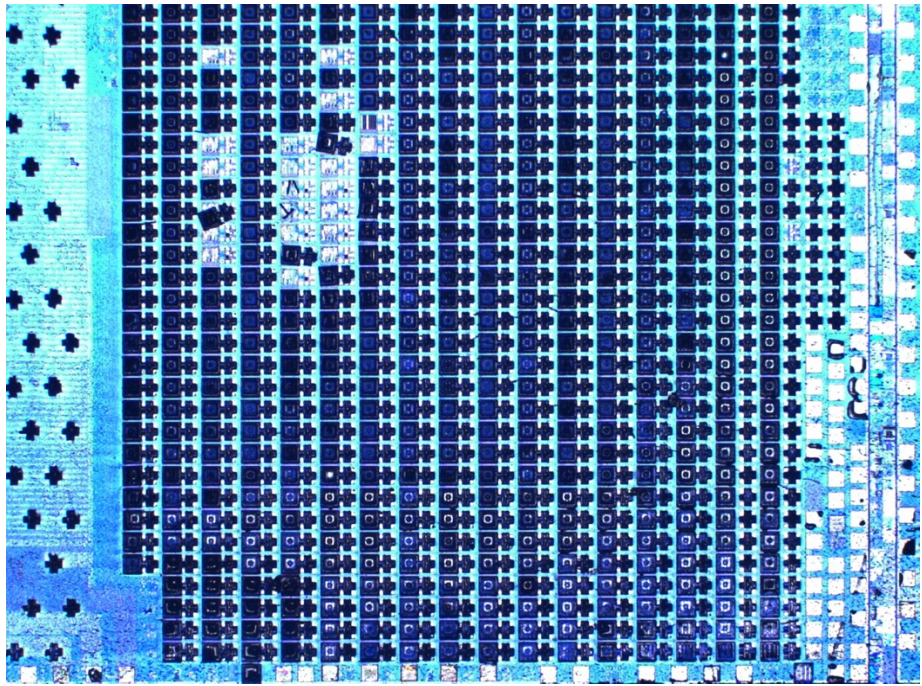

**Figure 21 Fully processed lablets in array.** Part of the array of lablets on one reticule after completed post-processing, prior to singulation. Some of the lablets have lost their laminate but yield is about 90% at this stage. The array of crosses on the left are test structures over the dock.

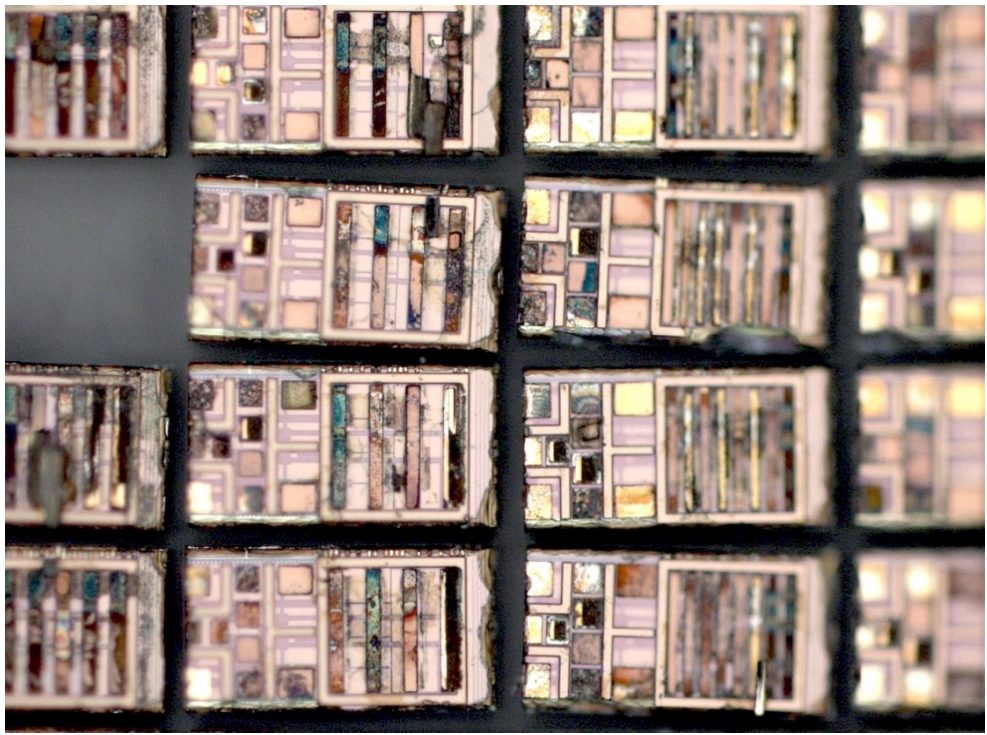

**Figure 22: Singulated lablets after full processing.** These examples show first results of singulation following post processing through all 4 post processing steps. In this case, the singulation removed the lablet wall and laminate structures and parts of the supercap coating. Note also that substantial chipping of structures is structurally critical in some cases.

## 5. Electronic viability of singulated lablets

The electronic functionality of thinned and singulated lablets was verified by making use of fine-tip probe needles. These probe needles have a tip diameter of a few micrometers to allow connecting to the lablet electrodes, including the supercap electrodes. In the measurement setup photographed in **Fig.**
**Figure 23**, the supercap connects to a voltage source of 1.8V, while the output electrodes of the lablet connect to an oscilloscope. The signal measured by the oscilloscope is depicted in **Fig.Figure 24.** As mentioned in the introduction, the lablet is able to generate signals based on a slow on-chip low power oscillator in combination with flip-flop based finite state machines.

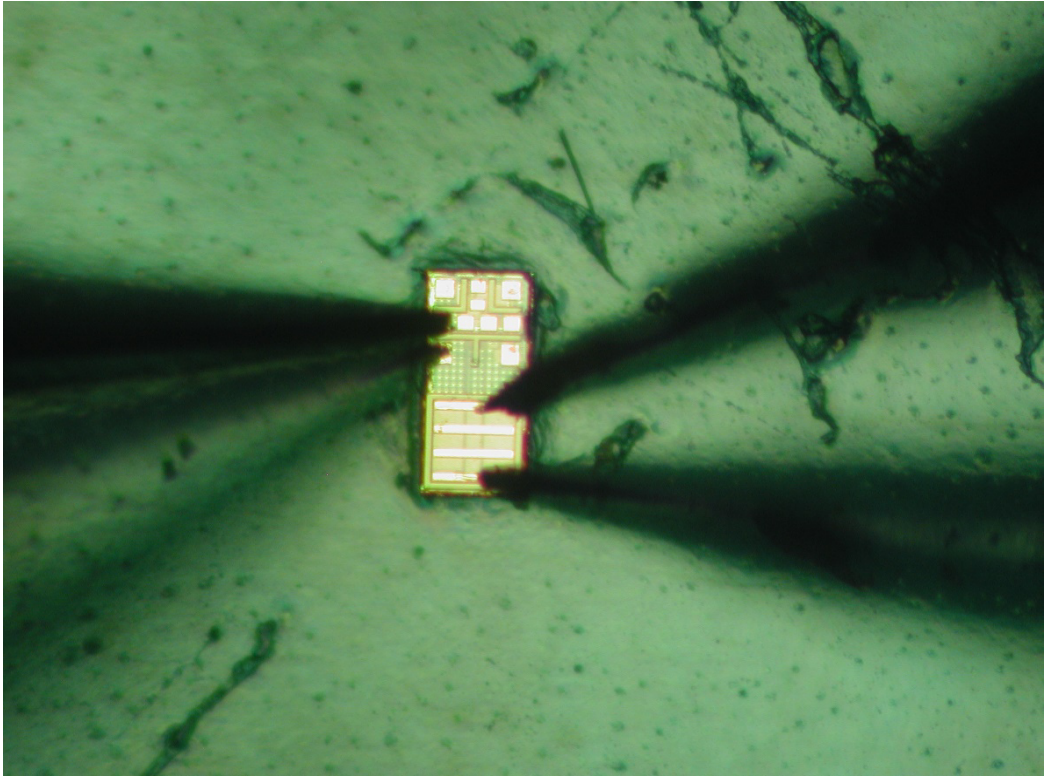

**Figure 23:** Microphotograph of singulated lablet with probe needles connected to it: The two probe needles on the right side are connected to the (uncoated) supercap electrodes to supply the electronics of the lablet with 1.8V. The two probe needles on the left side are connected to the output electrodes of the lablet and to an oscilloscope.

Using a similar approach, but with the addition of picolitre scale aqueous droplets, work was performed to extend this characterization to supercap powered coated lablets from which the laminate cover is absent. These experiments already show that the supercap coating does not disrupt lablet electronic functionality.

More extensive characterization of the electronic properties of lablets were performed on lablet arrays [10] but the current results demonstrate the autonomously clocked electrical viability of the singulated CMOS2 lablets.

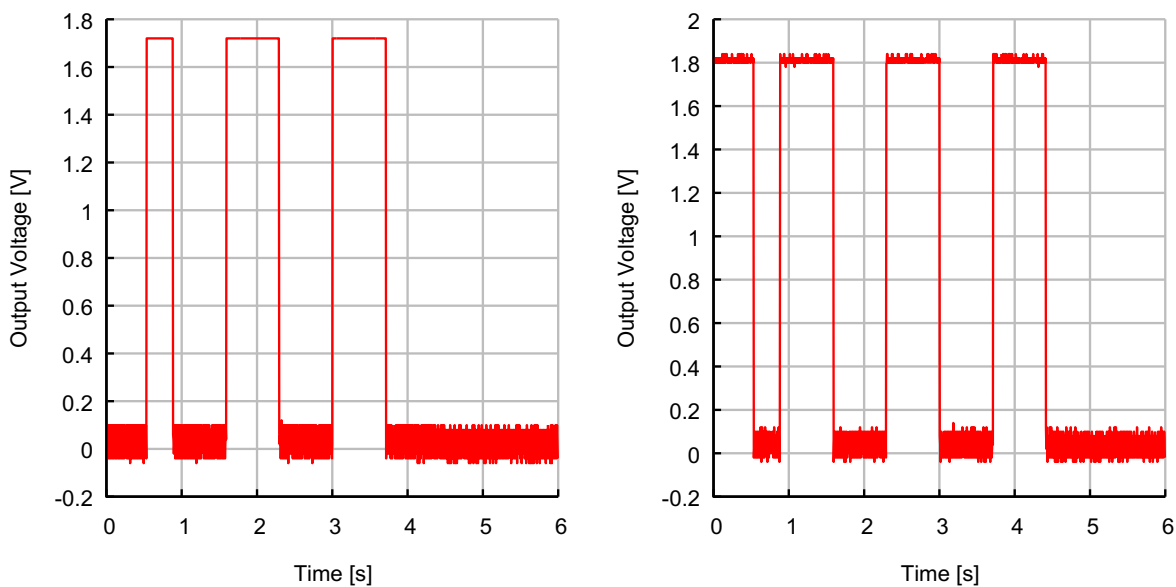

**Figure 24:** Measurement results of singulated lablet executing its own program. The two figures each correspond to one of the lablets output electrodes. The electrodes are either 0V, 1.8V or in high-impedance mode. In the high-impedance mode, no connection between electrodes and lablet circuity exists. Due to the input impedance of 1MΩ of the oscilloscope, high-impedance and 0V cannot be distinguished in this measurement.

## 6. Discussion and Conclusions

Lablets as presented here are only one possible realization of autonomous electrochemically active particles controlled by digital electronics. The authors are in doubt of their future widespread development and deployment as smart chemicals of the future, which will eventually permeate most areas of chemistry. It makes sense to produce a more endogenous linkup between electronic technology and the control of chemical reactions: especially since chemical reactions are primarily themselves rearrangement of electronic structures at the molecular scale. The main arguments against this approach are threefold: (i) the high energy cost of electronic switching (ii) the difficulties in controlling the electronic interface to chemistry and (iii) the need to scale down to sub-micron scales to ensure prolonged suspension and sufficient mobility of particles in solution. The general approach will be discussed in more detail elsewhere, but briefly let us discuss these three reservations here.

Purely molecular information processing, as in DNA Computing for example [25] can control chemistry in a more fine-grained manner with lower energy costs, and the authors have been early proponents of linking this approach with spatial microfluidics [26-28]. However, such systems are still limited in reliable dynamical and spatial switching and also require an interface to the external world for programming (usually via sequence synthesis) and readout (e.g. via sequencing). As the energy costs of gate transition decrease, these are moving towards molecular levels. Research in "zero-power" electronics and reversible computing are further narrowing this gap. Currently, various practical considerations (see introduction) have limited the access to highest resolution electronics that would have further decreased the cost of switching. Currently, the prime cost of power is less the switching of gates internally, but the switching of actuators because of their high double layer capacitance. The current approach overcomes this limitation using a supercapacitor as energy source. Further reduction in the size of microelectrodes can reduce the energy consumption in the interface towards molecular dimensions.

Electrochemistry is a well-developed science with an enormous range of redox reactions to choose from in programming chemical interfaces, both for actuation and sensing, but one where the influence of surface effects, and especially chemical absorption and other modifications of surface properties limit reproducibility unless strict electrode cleaning and polishing protocols are followed. Biology too makes extensive use of redox reactions in controlling processes at high integration density, and they are also involved in neural processing. While further research here is clearly warranted, the authors see a clear route to increased control of chemical via electronically produced potentials in future, and one where the reproducibility of integrated electronic protocols that can be deployed on lablets will facilitate the communicability and reproducibility of protocols between labs. Current lablets are restricted to providing a single digital voltage for actuation (this can be varied by charging level between deployments) and performing digital sensing of threshold conditions. Timing control of on-off signals can produce intermediate voltage levels in systems with large enough capacitance, but it is yet to demonstrate that this approach is effective in detailed electrochemical control on lablets. Similarly, actuator potentials can shift thresholds for sensing, in principle allowing the digital sensors to gain analog information.

The size of lablets is still a factor of five to ten larger than typical eukaryotic cells which can be suspended in solution with their nearly matched density, and for which electronic hybrid artificial analogs have been theoretically proposed [29]. The settling of suspensions above 1 µm in size is rapid when density differences occur, although stabilization by surfactants can occur, and indeed current lablet densities are between 2.5 and 3 times that of water. Whitesides has proposed magnetic levitation as a technique for suspending microparticles [30], and our project partner has developed an oil drop with ferromagnetic particles as an attachment approach for both levitating and externally manipulating the position of lablets [31]. We regard the levitation and autonomous motion of lablets as a next step in this research, after basic programmable chemical functionality has been achieved. This paper focusses on the constraints in the fabrication of the first lablets. Miniaturization towards the dimension of 1 µm, to allow the particles to be treated as real smart electronic colloids is conceivable over the next ten years with a concerted international technology effort.

Concerning the current fabrication of lablets, the approach that we have adopted raised galvanic processing to a central role in creating the coatings necessary for supercap and sensor functionality. While other masked approaches to oxide deposition could have produced a higher yield of lablets. in the case of $MnO_2$ for example, it was important for us to evaluate the power of differential galvanic processing in fabrication, because lablets themselves can and should eventually take control of specific electrode coatings via this technique in future. Difficulties in global interconnection of lablets (step 1) are real but can be overcome by integration of this step in CMOS fabrication or using an additional full planarization step prior to processing. This will also remove limitations in the galvanic processing with intervening protection barriers (step 2) since electrical contacting is then ensured at the wafer perimeter. Difficulties in encapsulation of electrolyte may be better overcome using ionogel electrolytes and replacing laminated sealing by spin coating, at the price of separating channel lid completion from supercap sealing (an extra process). Finally, the singulation of lablets is dependent on adhesion between interior layers being strong enough to withstand the necessary attachment and transfer of lablets between supports and the physical forces of thinning and sawing. An all-laser sawing process looks promising for further reducing these constraints. Additional testing of processes on less expensive substrates may ensure a higher yield of lablets even using the current approach in future.

Lablets are destined, when fully functional, to play a major role in chemical, biotechnological and, after a proper evaluation of risks, in medical technology. They not only provide an interface between digital electronic and chemical systems that can do down into microscopic dimensions, but they can serve as a platform for building autonomous chemically active systems *en route* to hybrid artificial cells. Other papers will deal with the investigation of lablet functionalities, possibilities, and applications. Ultimately, further miniaturized lablets can serve as a flexible interface to and control of the chemical environment both in synthetic and analytical contexts.

## Acknowledgements

This research is supported by the European Commission, EU FET Open MICREAgents Project # 318671. Authors at BioMIP are indebted to the excellent technical assistance of J. Bagheri-Maurer for chemical preparation and wafer handling. Wafer fabrication at TSMC was mediated with accommodation of special requests and rule checking by P. Malisse and J. Verherstraeten at IMEC. The authors wish to thank M. Richter and Atotech GmbH for complimentary electroless coating of wafers; H. and D. Bouwes and A. Kasjanow (iX Factory) for careful wafer cleaning and tests of alternative metal deposition strategies; S. Kaiser, M. Fleger and A. Lamhardt for extensive customer support during wafer processing at Temicon and for hosting our work on lamination in their clean rooms; H. Moritz at Silicet for accommodating our special requests and restrictions in the production of a galvanic wafer holder; and E. Eurich at Disco for his willingness to explore new options in wafer dicing for our wafers. Special thanks also go to N. Plumeré and W. Schumann at Electrochemistry, RUB for enabling AFM images and reference electrode construction; and to H. Hasenfratz for REM images of cut lablets, revealing the nature of various layers critical for galvanic processing. We wish to thank M. Zoshke and colleagues at FhG IZM, Berlin who evaluated the chip reassembly route to lablet planarization and fabrication, although this approach proved not competitive with wafer-scale fab, it taught us much in what was important in fabrication and alignment. Our thanks are also extended to K. Kallis TU Dortmund and A. Ludwig at RUB for their willingness to plan alternative fabrication routes with us. Finally, our thanks go to all the partners in the MICREAgents project who have been prepared to discuss our approach and born with us optimistically during this challenging fabrication. This manuscript was submitted in 2016 to the EU for external review as part of the final review of MICREAgents project, and has been reformatted here for ArXiv publication with minor updates to the references for those articles then in preparation, image completion for Fig. 14, and current first author affiliation.